\documentclass[twocolumn,superscriptaddress]{revtex4-1}

\usepackage{graphicx}
\usepackage{float}
\usepackage{epstopdf}
\usepackage{amsmath}
\usepackage{amssymb}
\usepackage{mdframed}
\usepackage[utf8]{inputenc}
\usepackage[dvipsnames]{xcolor}
\usepackage{tikz}

\usetikzlibrary{automata,positioning,arrows.meta}

\usepackage[caption = false]{subfig}

\usepackage{booktabs}

\usetikzlibrary{automata,positioning,arrows.meta}

\usepackage[caption = false]{subfig}

\makeatletter

\newcommand{\Rmnum}[1]{\expandafter\@slowromancap\romannumeral #1@}

\makeatother

\begin{document}

\title{Multiclass classification utilising an estimated algorithmic probability prior}

\author{Kamaludin Dingle}
\email{Correspondence: dinglek@caltech.edu}
\affiliation{Department of Computing and Mathematical Sciences, California Institute of Technology, USA}
\affiliation{CAMB,
Department of Mathematics and Natural Sciences, 
Gulf University for Science and Technology, Kuwait}

\author{Pau Batlle}
\affiliation{Department of Computing and Mathematical Sciences, California Institute of Technology, USA}

\author{Houman Owhadi}
\affiliation{Department of Computing and Mathematical Sciences, California Institute of Technology, USA}

\date{\today}

\begin{abstract}
\noindent
Methods of pattern recognition and machine learning are applied extensively in science, technology, and society. Hence, any advances in related theory may translate into large-scale impact. Here we explore how algorithmic information theory, especially algorithmic probability, may aid in a machine learning task. We study a
  multiclass supervised classification problem, namely learning the RNA molecule sequence-to-shape map, where the different possible shapes are taken to be the classes. 
The primary motivation for this work is a proof of concept example, where a concrete, well-motivated machine learning task can be aided by approximations to algorithmic probability. Our approach is based on directly estimating the class (i.e., \ shape) probabilities from shape complexities, and using the estimated probabilities as a prior in a Gaussian process learning problem. Naturally, with a large amount of training data, the prior has no significant influence on classification accuracy, but in the very small training data regime, we show that using the prior can substantially improve classification accuracy. To our knowledge, this work is one of the first to demonstrate how algorithmic probability can aid in a concrete, real-world, machine learning problem.\\ 

\vspace{0.35cm}
\end{abstract}


\maketitle
\twocolumngrid

\section{Introduction}
Machine learning (ML) is currently enjoying a great range of applications and rapid development. Applications of ML cover many areas of science, technology, industry, and society, and such applications are expected to significantly impact each of these areas. Hence there is interest in developing new methods to increase the accuracy or efficiency of pattern recognition, statistical learning, and ML tools.

One relatively unknown research area that may prove valuable to the ML community is \emph{algorithmic information theory} \cite{solomonoff1960preliminary,kolmogorov1965three,chaitin1975theory} (AIT), a field of computer science that studies the information content and complexity of discrete patterns, structures, and shapes. The amount of information required to describe or generate some pattern $x$ is quantified by its \emph{Kolmogorov complexity}, $K(x)$. Because AIT provides a general framework for defining and quantifying the notions of randomness and pattern --- which are fundamental to probability and statistics --- exploring ML, pattern recognition, and statistical learning from the perspective of AIT may lead to new theoretical and practical results.

Within AIT, an important concept invented by Solomonoff is \emph{algorithmic probability} \cite{solomonoff1960preliminary,solomonoff1997discovery,kirchherr1997miraculous,li2008introduction}, which examines the probability of outcomes via the information complexity of those outcomes. 
Building on Solomonoff,  Levin's \emph{coding theorem} \cite{levin1974laws} states that $P(x)=2^{-K(x)+O(1)}$ where $P(x)$ is the probability that $x$ appears from a general computation device on running a random program/algorithm. A deep result here is that probability estimates may be obtained directly from studying the complexity $K(x)$  of the pattern $x$, without referring to historical frequency statistical data, or the details of the mechanism that produced $x$. Hence algorithmic probability offers a fundamentally different view of probability estimation as compared to other common approaches. Herein lies the potential for new research directions and applications. 

Directly applying AIT and algorithmic probability to practical problems such as in ML is not at all straightforward, due to a number of reasons. These include the fact that Kolmogorov complexity is uncomputable, the theory is asymptotic with equations given only up to $O(1)$ terms, and the theory is developed in terms of universal Turing machines \cite{turing1936computable} (discussed below), which are not common in practical settings. Despite these and other issues, there are arguments that motivate the application of AIT in practical settings (see Appendix \ref{usingAIT}) in combination with various approximations and assumptions, and very many studies have applied AIT, including to make useful and verified predictions. Some examples are in physics  \cite{bennett1982thermodynamics,kolchinsky2020thermodynamic,zurek1989algorithmic,mueller2020law,avinery2019universal,martiniani2019quantifying}, biology   \cite{ferragina2007compression,adams2017formal,johnston2022symmetry,dingle2022fitness,dingle2022predicting}, networks \cite{zenil2014correlation,zenil2018review}, in addition to data analysis \cite{vitanyi2013similarity,cilibrasi2005clustering,zenil2019causal,zenil2011algorithmic,dingle2022note}, among several others \cite{li2008introduction,dingle2018input,dingle2020generic}.

AIT and algorithmic probability have been applied in ML, but these applications remain underdeveloped. To begin, both AIT and algorithmic probability were originally developed by Solomonoff in the 1960s as part of a research program into a general mathematical theory of induction and prediction. Hence the very origins of the area are closely connected to ML. Later, Schmidhuber \cite{schmidhuber1995discovering} wrote early work on AIT related to ML. More recently the works in refs. \cite{flood2020investigating,murena2020solving,schwartzman2021sgd,bernstein2021computing}  illustrate some applications of information arguments, data compression, and AIT to studies related to ML. Research in minimum description length (MDL) \cite{grunwald2019minimum}, a kind of computable applied AIT framework, has been utilized in statistics and data science quite commonly, e.g., \cite{dwivedi2020revisiting}, and most notably in model selection \cite{hansen2001model}. Valle-Perez et al.\ \cite{valle2018deep} (invoking results from \cite{dingle2018input}) have used AIT-based arguments to try to understand the surprising generalisability of deep neural networks.

Significantly, Hutter \cite{hutter2007universal} argues that algorithmic probability should form a key component of a general theory of prediction, and further that it provides a solution to the problem of choosing a Bayesian prior \cite{rathmanner2011philosophical} (although this philosophical claim is challenged \cite{neth2022dilemma}). Further, Hutter has developed a general theory of artificial intelligence, known as \emph{AIXI}, with algorithmic probability as a key component  \cite{hutter2004universal}. Although still largely a theoretical inquiry, numerical studies of the applicability of AIXI in toy models have been undertaken, via approximations of algorithmic probability \cite{veness2011monte,legg2013approximation}.
Relevant to our current study,
Zenil et al.\  \cite{zenil2019causal} have made conceptual contributions to the application of algorithmic probability to ML, and Zenil and collaborators have built on these to undertake other AIT and ML studies \cite{hernandez2021algorithmic,abrahao2021algorithmic} (discussed further below).

In the practical settings of biology, physics, and engineering, Dingle et al.\ \cite{dingle2018input} studied how approximations to algorithmic probability could be applied in  real-world input-output maps (see also \cite{dingle2020generic,johnston2022symmetry,dingle2022note,dingle2022predicting}). The main result was presenting an approximate upper bound $P(x)\lesssim 2^{-a\tilde{K}(x)-b}$, which was based on algorithmic probability but developed to be directly applicable in practical settings, unlike Levin's coding theorem. Here the term $\tilde{K}(x)$ is an estimate of the complexity of an output pattern $x$, computed via standard compression algorithms (see Appendix \ref{usingAIT}), and $P(x)$ is the probability that the pattern, sequence, or shape $x$ appears on a random sampling of inputs. This bound was used to help explain a preference for simplicity and symmetry in nature \cite{dingle2018input} (see also \cite{buchanan2018natural,johnston2022symmetry}).

Importantly, however, the upper bound can also be used to make \emph{a priori} probability predictions. What is especially noteworthy is that probability predictions can be made by directly examining the output patterns themselves, and computing their complexity values. This is in contrast to standard and well-known methods for estimating probabilities which rely on computing past frequency statistics, or utilizing detailed knowledge of the mechanism/map which produces the pattern.  The point is not that examining complexity estimates is more effective or accurate than frequency statistics or utilizing detailed knowledge of the map, but that in situations where these two standard approaches are not applicable (e.g., when no data is available, or only very limited data), the complexity estimation approach provides a completely different method which may be used to make informative predictions, even if not always very accurate.
 
Motivated by the success of these algorithmic probability-based predictions, and the potential for advancements in ML via AIT, here we study a concrete ML classification problem that combines a Gaussian process learning algorithm with the practical algorithmic probability estimates of Dingle et al.\  \cite{dingle2018input}. Specifically, we study a multiclass supervised classification problem in which an RNA sequence-to-shape map is learned from data, where the different possible RNA shapes are taken to be the classes. With large amounts of training data, the choice of prior becomes irrelevant to classification accuracy, but in cases where only small training data sets are available, a `good' prior can in theory improve the accuracy of classification predictions. In our context, the `best' prior would be to use the true class probabilities (i.e.\ the RNA shape probabilities), but these class probabilities are typically not known a priori. However, the mentioned practical algorithmic probability estimates provide a way to directly estimate these class probabilities, via measuring the complexities of each shape. The main aim of this current investigation is to numerically study if multiclass classification accuracy can be improved in small training data regimes, while using the algorithmic probability estimates of ref.\ \cite{dingle2018input} to predict the class probabilities.

Note that the recent study of Hernández-Orozco et al.\ \cite{hernandez2021algorithmic} is related to ours in that algorithmic probability is applied to ML problems, including a classification problem. However, our study differs substantially, including in terms of the method, the application area, and fundamentally how the algorithmic probability is incorporated.

 \section{Theory and problem set-up}

 \subsection{Kolmogorov complexity}
We give some brief background to AIT for the sake of completeness.

Developed within theoretical computer science, \emph{algorithmic information theory} \cite{solomonoff1960preliminary,kolmogorov1965three,chaitin1975theory} (AIT)  connects computation, computability theory and information theory. The central quantity of AIT is \emph{Kolmogorov complexity}, $K(x)$, which measures the complexity of an individual object $x$ as the amount of information required to describe or generate $x$.  More formally, the Kolmogorov complexity $K_U(x)$ of a string $x$ with respect to a universal Turing machine \cite{turing1936computable} (UTM) $U$,  is defined \cite{solomonoff1960preliminary,kolmogorov1965three,chaitin1975theory} as
\begin{equation}
K_U(x) = \min_{p}\{|p|: U(p)=x\}
\end{equation}
where $p$ is a binary program for a prefix (optimal) UTM $U$, and $|p|$ indicates the length of the (halting) program $p$ in bits.  Due to the invariance theorem \cite{li2008introduction} for any two optimal UTMs $U$ and $V$, $K_U(x) = K_V(x)+O(1)$ so that the complexity of $x$ is independent of the machine, up to additive constants. Hence  we conventionally drop the subscript $U$ in $K_U(x)$, and speak of `the' Kolmogorov complexity $K(x)$. Despite being a fairly intuitive quantity and fundamentally just a data compression measure, $K(x)$ is uncomputable, meaning that there  cannot exist a general algorithm that for any arbitrary string returns the value of $K(x)$.
Informally, $K(x)$ can be defined as the length of a shortest program that produces $x$, or simply as the size in bits of the compressed version of $x$. If $x$ contains repeating patterns like $x=1010101010101010$ then it is easy to compress, and hence $K(x)$ will be small. On the other hand, a randomly generated bit string of length $n$ is highly unlikely to contain any significant patterns, and hence can only be described via specifying each bit separately without any compression, so that $K(x)\approx n$ bits. $K(x)$ is also known as  \emph{descriptional complexity}, \emph{algorithmic complexity}, and \emph{program-size complexity}, each of which highlight the idea that $K(x)$ measures the amount of information required to describe or generate $x$ precisely and unambiguously.

More details and technicalities can be found in standard AIT references \cite{li2008introduction,calude2002information,gacs1988lecture,shen2022kolmogorov}.

\subsection{Algorithmic probability}

In AIT,  Levin's \cite{levin1974laws} coding theorem establishes a fundamental connection between $K(x)$ and probability predictions. Building on Solomonoff's  discovery of \emph{algorithmic probability} \cite{solomonoff1960preliminary,solomonoff1997discovery}, Levin's coding theorem \cite{levin1974laws} states that
\begin{equation}
P(x) = 2^{-K(x)+O(1)}\label{eq:CTHM}
\end{equation}
where $P(x)$ is the probability that (prefix optimal) UTM $U$ generates output string $x$ on being fed random bits as a program. Thus, high-complexity outputs have exponentially low probability, and simple outputs must have high probability. $P(x)$ is also known as the \emph{algorithmic probability} of $x$.

\subsection{Simplicity bias}
The coding theorem stated in Eq.\ (\ref{eq:CTHM}), as well as many other AIT results cannot be straightforwardly applied to typical natural systems of interest in engineering and sciences, due to the fact that: (1) Kolmogorov complexity is uncomputable and so cannot be calculated exactly; (2) the theory is asymptotic,  valid only up to $O(1)$ terms; (3) the theory is largely based on UTMs, which are seldom present in nature; and (4) the coding theorem assumes infinite purely uniform random programs,  which do not exist in nature. 
Despite these points, several lines of reasoning motivate using AIT to make predictions while being aware of the limitations of this practice (Appendix \ref{usingAIT}).  We call this kind of theoretical work `AIT-inspired' arguments .

Adopting the methodology of AIT-inspired arguments, Dingle et al.\ \cite{dingle2018input} studied coding theorem-like behaviour and algorithmic probability for (computable) real-world input-output maps.  This led to their observation of  \emph{simplicity bias} (SB), governed by the equation 
\begin{equation}
P(x)\lesssim 2^{-a\tilde{K}(x)-b}\label{eq:SB}
\end{equation}
where $P(x)$ is the (computable) probability of observing output $x$ on random choice of inputs, and $\tilde{K}(x)$ is the approximate Kolmogorov complexity of the output $x$. In words, SB means complex outputs from input-output maps have lower probabilities, and high probability outputs are simpler. The constants $a>0$ and $b$ can be fit with little sampling and often the default values of $a=1$ and $b=0$ work well, without the need to fit to data \cite{dingle2018input}. 

Eq.\ (\ref{eq:SB}) is based on estimating the compressed size of a pattern $x$, and hence the representation of $x$ can affect the complexity estimate $\tilde{K}(x)$. As an extreme example, if $x$ were encrypted using a pseudo random number, then even if $x$ were compressible in its original form, then the encrypted version would be deemed incompressible by most compression algorithms, and hence assigned a high complexity. Despite this theoretical possibility, the empirical success of SB studies demonstrate that this possibility does not invalidate the usefulness of predictions based on SB.

Examples of systems exhibiting SB are wide-ranging, and include molecular shapes such as protein structures and RNA \cite{johnston2022symmetry}, outputs from finite state machines \cite{dingle2020generic}, as well as models of financial market time series and ODE systems \cite{dingle2018input}, among others. 
A full understanding of exactly which systems will, and will not, show SB is still lacking, but the phenomenon is expected to appear in a wide class of input-output maps, under fairly general conditions (Appendix \ref{ap:conditionsSB}).



Although Eq.\ (\ref{eq:SB}) is only an upper bound, it is tight with high probability \cite{dingle2020generic}, i.e., randomly chosen inputs are likely to map to outputs which are close to the bound, so that $P(x)\approx 2^{-a\tilde{K}(x)-b}$. However the existence of low-complexity, low-probability outputs \cite{dingle2020generic,alaskandarani2022low} means that for many $x$ we may $P(x)\ll 2^{-a\tilde{K}(x)-b}$.

Using the default values $a=1$ and $b=0$, Eq.\ (\ref{eq:SB}) can be used to predict the probability of patterns or shapes by making the approximation that $P(x)$ is actually equal to the upper bound. Hence the prediction ${\hat P}(x)$ for the true probability $P(x)$ is
\begin{equation}
{\hat P}(x) = 2^{-\tilde{K}(x)}\label{eq:SB_pred}
\end{equation}
Note that in practice, the ${\hat P}(x)$ values often do not sum to 1 (discussed later). Hence using Eq.\ (\ref{eq:SB_pred}), probability predictions can be made by directly estimating the complexity $\tilde{K}(x)$ of the pattern $x$, while being aware that the accuracy of the predictions vary considerably and are sometimes quite poor. Despite this, we consider it is better to have predictions of varying accuracy than no predictions at all: the major advantage of Eq.\ (\ref{eq:SB_pred}) is that it can be used in situations where typical common methods for probability predictions fail, such as cases where no statistical frequency data are available to make probability predictions. 


 \subsection{Gaussian processes}
 
 Gaussian processes are a common and powerful method in ML \cite{bishop2006pattern,williams2006gaussian}. We will study a multiclass supervised learning classification problem. Although a classification problem, we will use Gaussian process regression to solve the problem by learning a real-valued vector which indicates how likely a given test sample is to belong to each of the multiple classes. 
 We use a one-hot vector to encode the class. A one-hot vector is a vector of length $C$,  with $C$ being the number of classes, where one element of the vector is 1, and all others are 0. The location of the 1 indicates the class. The Gaussian process regression algorithm predicts the real-valued vector, 
 and the class prediction is simply the vector element with the highest score.
For example, if $C=5$ and the predicted vector for some given test sample is (0.01, 0.00, 0.40, 0.10, 0.20) then the prediction for the test sample is that it belongs to the 3rd class. Note that it is not necessary for the real-valued vector to sum to 1. 
 
The mean of a Gaussian process is typically set to a zero vector in ML applications. However, if prior information is available for the expected value of each variable, then this prior knowledge can be expressed via the mean. The central idea of this study is to use $\hat{P}(x)$ in Eq. (\ref{eq:SB_pred}) as the prior mean of the Gaussian process. 

 We write a Gaussian process as
 \begin{equation}
 f({\bf x}) \sim \mathcal{GP}(m({\bf x}), k({\bf x}, {\bf x'}))
 \end{equation}
 where $m({\bf x})$ is the mean function, and $k({\bf x}, {\bf x'})$ is the covariance function. After having chosen a mean and covariance function and having observed a dataset $\{({\bf x_i }, {\bf y_i})\}_{i= 1}^n$, where ${ \bf y_i}$ are potentially noisy observations of $f({\bf x_i})$, the conditional expectation of the Gaussian process is the function
 \begin{equation}
 {\bar f} ({\bf x}) = m({\bf x}) + K({\bf x},X)(K(X,X)+\lambda Id)^{-1}(Y -m(X))
 \end{equation}
 where $K(X,X)$ is a matrix with $K(X,X)_{i,j} =  k({\bf x_i}, {\bf x_j})$, and $K({\bf x},X)$ is a vector with $K({\bf x},X)_i = K({\bf x},{\bf x_i})$. The same stacking notation applies to $Y$ (stacking of the ${\bf y_i}$) and $m(X)$ (stacking of $m({\bf x_i})$). The term $\lambda > 0$ can be used to model noisy observations, in this work we allow the \emph{Scikit-learn} \cite{scikit-learn} Gaussian process \texttt{fit()} to determine the value that best explains the data (using maximum likelihood estimation). 
 In words, if we have a prior for the class probabilities, we can incorporate this information via the mean, then use Gaussian process regression to learn the difference between the mean and the vector values of the class, and then add the mean back again at the end before converting the real-valued vector into a binary vector to be used for the test sample class prediction.

\subsection{RNA sequence-to-shape map}
 
RNA are important and versatile biomolecules that act in organisms as information carriers, catalysts, and structural components, among other roles. Similar to DNA, RNA is comprised of a sequence of nucleotides (``letters''), which can contain four possible nucleobases, A, U, C, and G. RNA molecules typically fold up into well-defined secondary structures, which denote bonding patterns in the molecule. 
The structure is determined by the underlying sequence, so that different sequences of nucleotides typically yield different structures. 
At the same time, the sequence-to-structure map is highly redundant, with many different sequences adopting the same structure \cite{schuster1994sequences}. Because there are four possible letters per base, there are $4^L$ different possible sequences of length $L$. The number of different possible structures scales as $1.76^L$ \cite{dingle2015structure}, which is much less than the number of possible sequences, but still exponential in $L$. RNA structure has been very well studied in biophysics and computational biology because it is a biologically relevant system, but at the same time, fast and accurate computational algorithms exist for predicting a structure from an arbitrary sequence \cite{lorenz2011viennarna,liu2022expertrna}. These algorithms are often based on thermodynamic principles, with a sequence's minimum free energy structure taken as `the' structure for a given sequence. Other structure prediction approaches also exist, such as those based on evolutionary conserved sequence patterns, but we will not consider these here.

RNA secondary structures can be coarse-grained to abstract shapes, which are intended to represent the overall shape of the molecule while ignoring finer details of the structure \cite{giegerich2004abstract,janssen2014rna}, such as the lengths of loops and stems. These simplified abstract RNA shapes are used in bioinformatics, and molecular evolution studies \cite{dingle2022phenotype,martin2021insertions,johnston2022symmetry,ghaddar2022random}. In this work, we will examine the map from RNA sequences to abstract shapes (specifically, using `level 5' abstraction  \cite{giegerich2004abstract,janssen2014rna}).

\begin{figure}
\subfloat[]{\includegraphics[width=9cm]{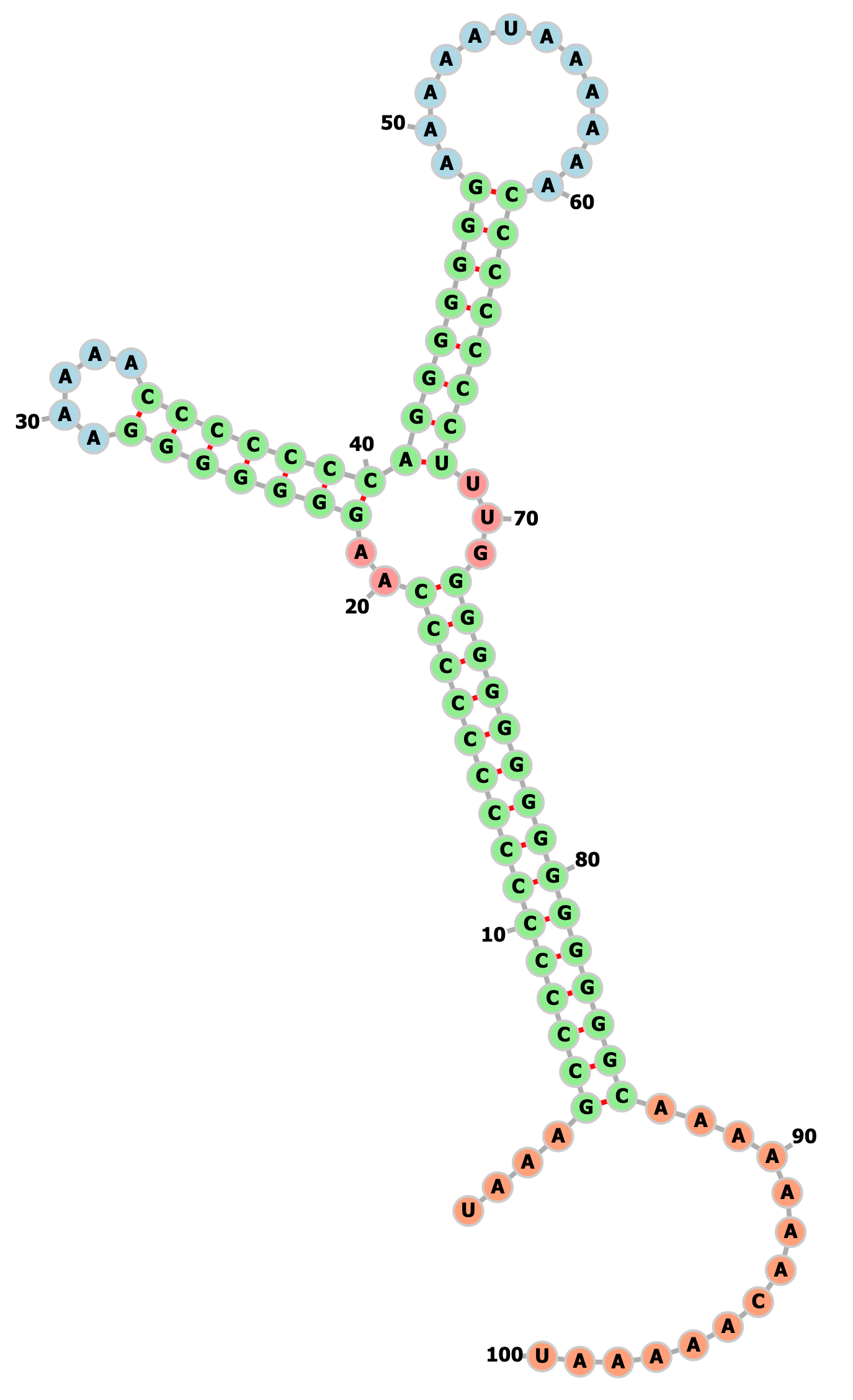}}
\caption{\textbf{Example RNA  structure}. A sequence of length $L=$ 100 letters (nucleotides) was chosen for illustration purposes. This sequence adopts the structure shown. Because the structure is formed of two stem and loop motifs, within a larger third stem, it has abstract shape [[][]].}
\label{fig:RNA}
\end{figure}

In Figure \ref{fig:RNA}, an example length $L=$ 100 nucleotide sequence is displayed in its   folded form, that is, the minimum free energy predicted secondary structure. 
The computational structure prediction was implemented using the Vienna RNA package \cite{lorenz2011viennarna}, and the illustration was drawn using \texttt{rna.tbi.univie.ac.at/forna}. The nucleotide bases are numbered 1 to 100 for illustration purposes. As is apparent, some of the bases are bonded to other bases, and these bonded pairs are coloured in green. Other bases are unbonded, and these are coloured in red or blue.
By varying the sequence, many different possible structures and shapes are possible. 
Because the structure displayed is essentially just two stem and loop motifs (bases 22---40 and 41---68), which both sit within a larger stem (formed of bases 5---19 and 72---86), the structure can be abstracted to just [[][]], where the pair of brackets ``[]'' indicates a bonded stem.

To avoid potential confusion, we highlight that we use the word ``structure'' for the full bonding pattern of an RNA (e.g., depicted in Figure \ref{fig:RNA}), and we use the word ``shape'' to mean the abstract square bracket representation, e.g. [[][]]. Therefore ``structure'' and ``shape'' are not synonymous in this work.

To make nucleotide sequences more amenable to machine learning methods, we will convert nucleotide sequences of letters into numerical vectors as follows: we replace the letter A with the string 1000, the letter U (or T) with 0100, C with 0010, and G with 0001. Subsequently, the resulting binary string is converted into a vector. In this manner, a length $L$ sequence of letters becomes a length $4L$ binary vector. For example, the sequence AGC would become the vector (1,0,0,0,0,0,0,1,0,0,1,0). The Hamming distance between two sequences of letters can therefore be computed by finding the Hamming distance between two such binary vectors, and multiplying by 0.5.

Although there already exist computational methods that can predict these RNA shapes directly from RNA sequences using biophysics principles, as a well-motivated and practical case study in which to explore algorithmic probability in ML,
we will study the problem of learning the sequence-to-shape map for RNA given some training data of sequence-shape pairs. In ML terminology, this is a multiclass supervised classification problem, where given some examples of sequence-shape pairs, we try to predict the class (shape) of new unseen RNA sequences. 

Finally, note that we will be attempting to learn the sequence-shape map from data generated by employing the Vienna RNA computational prediction algorithm. This algorithm is designed to replicate natural RNA folding patterns, but is of course, only a model of that process. Hence we are employing a surrogate model, for convenience. This qualification of our method is not very significant, however, because the primary aim of this work is a proof of concept example of how a prior based on algorithmic probability can be used in ML. The Vienna RNA package is convenient and hence employed, but any (large) collection of sequence-shape pairs and prediction algorithm
could in principle, be used in combination with our proposed prior.


  \begin{figure}
\subfloat[]{\includegraphics[width=9cm]{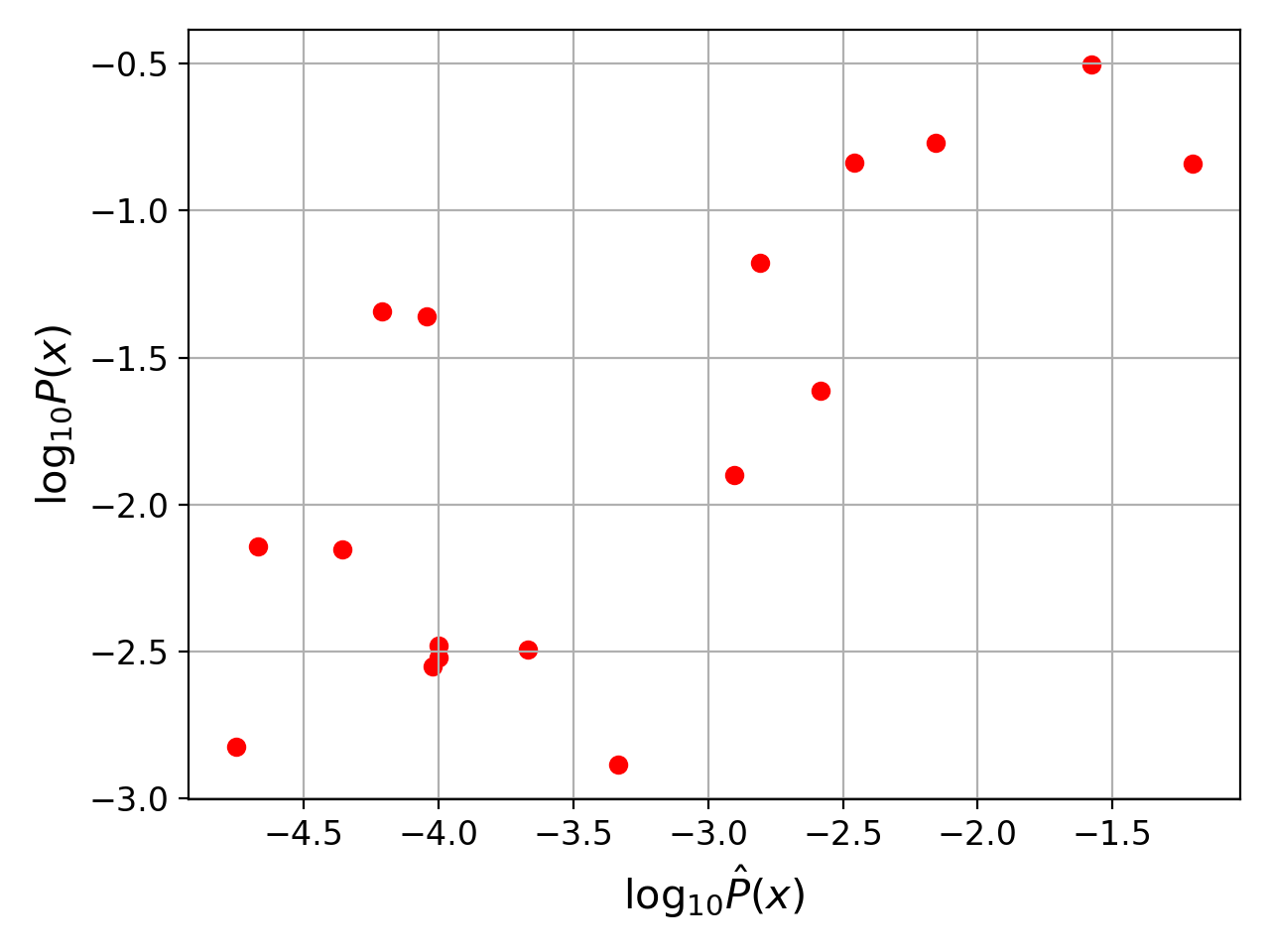}}
\caption{\textbf{Predicting class probabilities}. Predicted class probabilities  $\hat{P}(x)$ using Eq.\ (\ref{eq:SB_pred}) plotted against true class probabilities $P(x)$ for RNA shapes $L=100$. There are 17 shapes (classes). The linear correlation is quite strong at $r=0.74$.}
\label{fig:correlation}
\end{figure}

 \section{Numerical experiments}
 
\subsection{Classification method details}

\subsubsection{Training and testing data}
We now describe the implementation details for the classification problem:
 We used a (pseudo) random number generator to generate 10,000 RNA sequences of length $L=100$, with each letter A, U, C, G having equal probability. We computed the corresponding RNA secondary structures using the RNA Vienna folding package \cite{lorenz2011viennarna}, and found the corresponding abstract shapes (at level 5 abstraction) with the algorithm given in  \cite{giegerich2004abstract,janssen2014rna}.  We discarded any sequences whose corresponding shapes had less than 10 associated sequences. The reason for this was to prevent having only a trivial amount of training data for a given shape. 

In the generated data, only 17 unique shapes appeared (recall that many different sequences can share the same shape). The shapes were [[[][]][]], [[][[][]]], [[][][][]], [[][][]], [[][][]][], [[][]], [[][]][], [[][]][][], [], [][[][][]], [][[][]], [][[][]][], [][], [][][[][]], [][][], [][][][], and [][][][][]. These 17 shapes are taken as the 17 classes for the multiclass classification problem (see Appendix \ref{app:classdefinition}). 

By counting the frequency of each shape in the full set of 10,000 samples, we computed the `true' class probabilities $P(x)$ for each shape/class. The probabilities are [0.0028, 0.003, 0.0126, 0.0663, 0.0072, 0.1463, 0.0438, 0.0032, 0.1444, 0.007, 0.0454, 0.0015, 0.3141, 0.0033, 0.1702, 0.0243, 0.0013].
 
Of the 10,000 generated samples, 1000 were set aside as the testing set (but see Appendix \ref{app:classdefinition}). The training data samples were chosen from the remaining $\sim$9,000 samples. 

Because the influence of the prior is greater for small training data sets, we explore classification accuracy for various small training data set sizes.  To begin, we chose only 1 sequence (at random) for each shape, yielding a training set of only 17 samples (one per class), and then computed classification accuracy scores using the test set. Next, we chose 2 sequences for each shape and computed classification accuracy. Proceeding in this manner, we gradually increased the number of samples per shape using these values:  1, 2, 3, 5, 10, 20, 40, 80, 160, 320, 640, 1280, 2560, and 5120. 

Note that the different shapes are  known to have order-of-magnitude differences in the numbers of sequences per shape \cite{schuster1994sequences,dingle2022phenotype}, hence as the number of samples per shape increased some shapes did not have the requisite number of sequences. For example, out of the training samples, if a shape only had 19 sequences that underlie it, obviously it is not possible to select 20 or more sequences for that shape. In these cases, we merely took whatever sequences were available. Therefore the sizes of the training data did not always proceed as multiples of 17.

The way the training data is constructed is an important consideration. We purposely did not construct the training data from randomly chosen sequences, for two reasons: First, it is  common that ML tasks are performed using non-random data sets \cite{liu2021inference,rafei2022robust}, which can arise for many reasons such as non-random data collection processes in voting preference polls, or non-random database construction in bioinformatics arising from a particular interest in certain diseases and associated molecules. Second, the purpose of this study is specifically to investigate if estimating class probabilities via algorithmic probability arguments will improve classification accuracy, whereas if random sequences are used for the training data then the training data itself would  already contain good estimates of class probabilities. Hence there would be little need for independent estimates of class probabilities via complexity arguments, which may be less accurate than those directly obtained from the training data.

As a supplementary example, in addition to the $L=$ 100 data just described, we also study RNA data of $L=$ 60. See Appendix \ref{app:L60} for more details.
 
 \subsubsection{Gaussian process implementation}
 From considering the biophysics of RNA folding, we expect sequence similarity (i.e., Hamming distance) to be a main factor in predicting whether two sequences adopt the same shape (although it is possible that two sequences with high similarity have very different shapes \cite{schuster1994sequences}). Hence choosing a distance-based kernel is  reasonable for  the Gaussian process algorithm. There are many possible distance-based kernels, and of these, we chose the Gaussian kernel, also known as the radial basis function (RBF), or squared exponential. The reason for this choice of kernel is that the Gaussian is a universal kernel, and is often taken as a default. While a more careful choice of kernel will likely improve classification accuracy,  this current  study is not specifically concerned with kernel selection (but see refs.\ \cite{owhadi2019kernel,owhadi2021kernel} for an approach to learning kernels from data).
 
To implement the Gaussian process classification, we used the popular Python ML library \emph{Scikit-learn} \cite{scikit-learn}, including the \texttt{fit()} method for learning the kernel length scale,  with $1/C$ multiplied by the mean distance between sequences as a starting point (recall that $C$ is the number of classes). 


 \subsubsection{Complexity estimation and probability prediction}
 
 The complexity of the 17 shapes was estimated by converting each shape to a binary string, by replacing [ with a 0, and ] with a 1. For example, the shape [][][] becomes 010101. Subsequently, these binary string complexity values $\tilde{K}(x)$ were estimated based on a slightly modified Lempel-Ziv complexity \cite{lempel1976complexity} measure, as used earlier \cite{dingle2018input,dingle2020generic} (see Appendix \ref{app_complexity}).
 
 The predicted class probabilities $\hat{P}(x)$ can be then calculated using these complexity values and Eq.\ (\ref{eq:SB_pred}). However, because the complexity values $\tilde{K}(x)$ are noisy imprecise complexity estimates, and $\hat{P}(x)$ is exponentially dependent on these estimates, we additionally `smoothed' the probability estimates $2^{-\tilde{K}(x)}$ by geometric averaging subsets of sorted probabilities (Appendix \ref{app_avprb}).  The resulting predicted class probabilities $\hat{P}(x)$ are [0.0001, 0.0001, 0.0012, 0.0016, 0.00002, 0.0035, 0.0001, 0.0002, 0.0625, 0.00004, 0.0001, 0.00002, 0.0263, 0.0001, 0.007, 0.0026, 0.0005]. Note that the sum of $\hat{P}(x)$ is 0.11, which is less than 1, but this is not a problem because there is no need for the prior expressed as the mean to sum to 1 (discussed more below).
 
 \subsection{Results}
 \subsubsection{$\hat{P}(x)$ vs $P(x)$}
 
In Figure \ref{fig:correlation} we plot the predicted class (shape) probabilities $\hat{P}(x)$ against the true class probabilities $P(x)$. (Because 10,000 samples were used, and some probabilities are $\approx10^{-4}$, there are likely some statistical sampling errors in the `true' probability values). Because of the order-of-magnitude variation in probabilities, the values are plotted as logarithms. 

The Pearson linear correlation is quite strong with $r=$ 0.74 and statistically significant because the $p$-value = 0.0006 $\ll$0.05 (assuming 5\% as a significance threshold). This correlation is quite impressive, given that it is based on quite abstract AIT-inspired arguments, the predictions coming directly from looking at the shapes themselves. For the supplementary $L=$ 60 data, the correlation is very high, with $r =$ 0.97 (Appendix \ref{app:L60}).

Although the predicted values have high linear correlation for both $L=$ 100 and $L=$ 60, these predictions tend to underestimate the true values by a roughly constant factor (consistent with the fact that the sum of $\hat{P}(x)$ is less than the sum of $P(x)$). This underestimation is not a great problem. There is no requirement for the prior, incorporated via the mean of the Gaussian, to sum to 1. Trivially, the mean of a Gaussian process is typically set to zero, which does not sum to 1. 

More significantly, it maybe be more prudent to impose a mean (prior) with smaller norm, unless there is great confidence in the quality of the prior. Hence the fact that the sum of $\hat{P}(x)$ is less than 1 may in fact be desirable. In the case that a prior is generated from somewhat loose complexity arguments, one should not be overly confident about the prior. Following this reasoning, in practice, if very uncertain about the quality of the prior it may be more scrupulous to use $\alpha\hat{P}(x)$, for $\alpha\leq 1$, to reduce the influence of the prior, and the value of $\alpha$ could be tuned precisely via cross-validation. We leave this to future work to explore.

\begin{figure*}
\subfloat[]{\includegraphics[width=9cm]{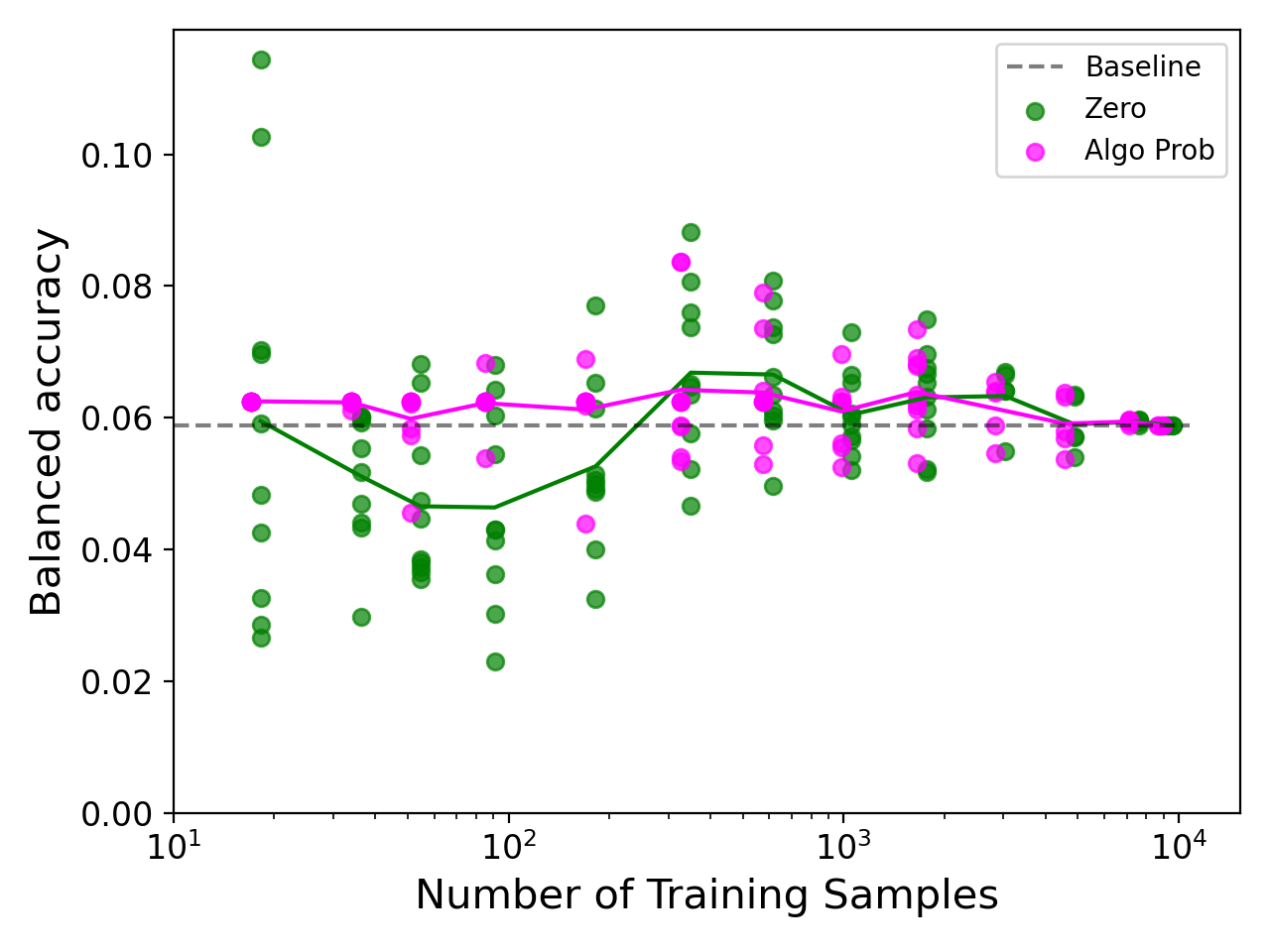}}
\subfloat[]{\includegraphics[width=9cm]{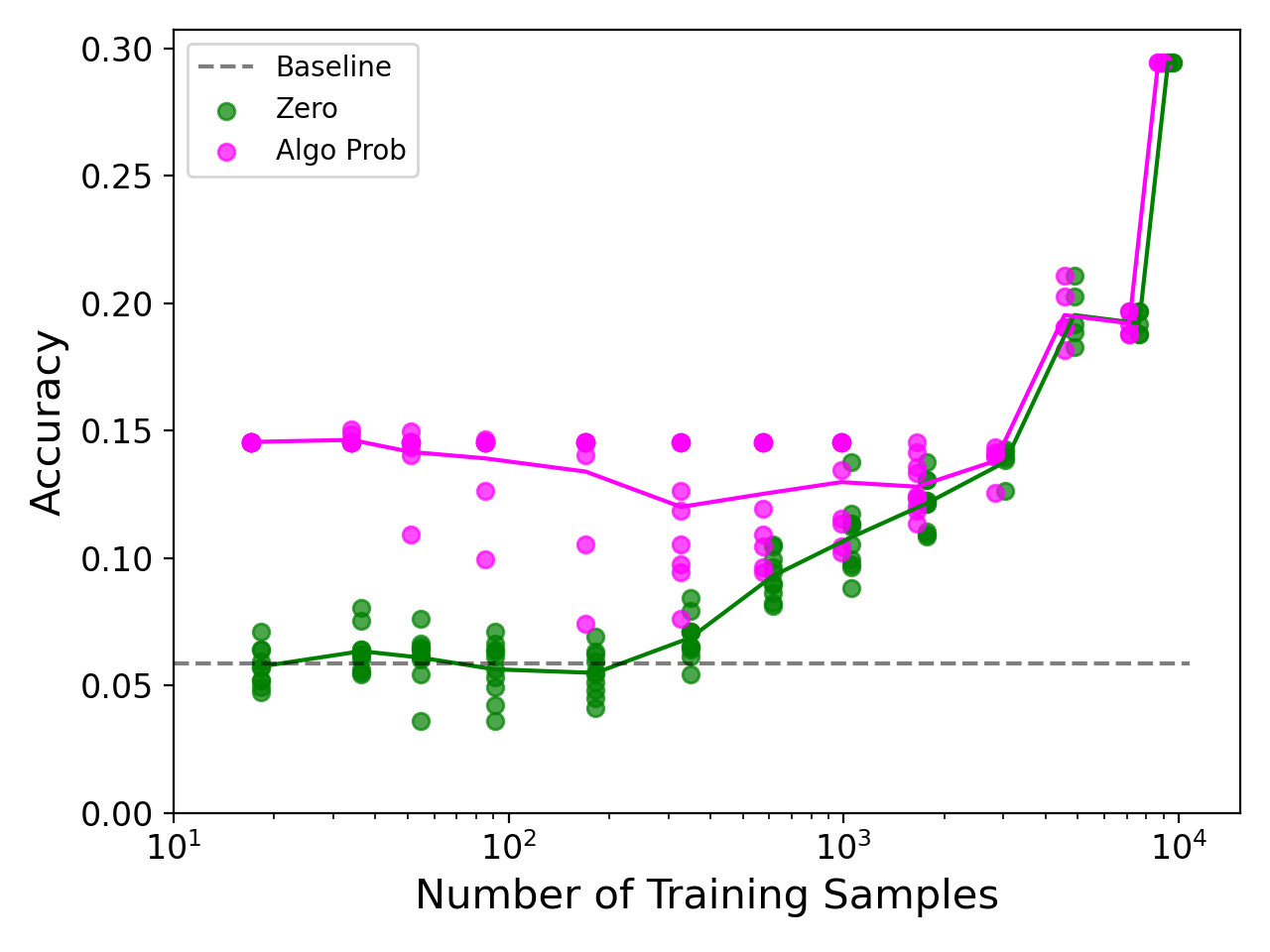}}\\
\caption{\textbf{Multiclass classification accuracy using different priors}. Comparing the accuracy for a 17-class supervised learning classification problem with a zero mean prior (green), and the algorithmic probability-based prior $\hat{P}(x)$ of Eq.\ (\ref{eq:SB_pred}) (pink) for different training sample sizes. Panel (a) shows mean balanced accuracy scores when equally weighting each class; (b) shows mean accuracy, i.e., simply the fraction of correct classifications. In (a), both priors yield similar accuracy scores, but in (b), the  $\hat{P}(x)$ prior outperforms the zero mean prior for smaller sample sizes. Note that a small multiplicative shift has been applied to separate the green and pink data, for the purposes of visualisation only.}
\label{fig:accuracy}
\end{figure*}

 \subsubsection{Classification accuracy}
 
 Having observed that we can make reasonably accurate class probability predictions --- suggesting that $\hat{P}(x)$ is likely a good prior --- we now turn to the main goal of this study, which is to see if classification accuracy can be improved by using $\hat{P}(x)$ as a prior. For comparison, we also computed the classification accuracy for the same training and testing datasets, but while using the standard zero mean Gaussian process. This zero mean benchmark is what we hope to improve on. Also for comparison, the baseline accuracy given by (1/$C$) is shown. This baseline is the accuracy achievable by simply randomly assigning sequences to testing shapes.



In the multiclass classification problem, classification accuracy is measured in two ways. Firstly by the  \emph{balanced accuracy} using \emph{Scikit-learn's} inbuilt method \texttt{metrics.balanced\_accuracy\_score}, and secondly by \emph{accuracy} using  \texttt{metrics.accuracy\_score}. The balanced accuracy is intended to give equal weighting to each class, and the plain accuracy simply measures the fraction of correct classifications. We include the balanced accuracy for completeness, while the plain accuracy is the more relevant measure for this RNA problem. While there may be certain biological settings in which it is important to accurately classify rare or specific RNA, and in these settings, balanced accuracy (or some other bespoke metric) may be more relevant, in general, simply counting the number of correct classifications is appropriate. Note that due to the strongly non-uniform probability distribution over RNA shapes, the balanced accuracy and the accuracy may be quantitatively quite different in this setting.


Turning now to the results of our classification numerical experiments, 
Figure \ref{fig:accuracy}(a) shows the classification balanced accuracy plotted against different training data sizes. It is apparent that the two priors achieve essentially the same balanced accuracy, except for very small training data sizes, for which the algorithmic probability-based prior shows a modest improvement. Hence using the algorithmic probability-based prior  does not improve performance in a substantial way for this metric. Neither method improves on the baseline accuracy given by $1/C$, even for training data sizes of up to $\sim$9,000. This observation can be rationalised by recalling that the training data are  100-dimensional due to the 100 nucleotide letters, hence very high dimensional, and so it is not remarkable that large training data set sizes are required for higher accuracy classifications. Additionally, the balanced accuracy gives equal weighting to all classes, and due to the fact that most of the classes have very low probability, even within $\sim$9,000 training samples, many of the classes have very few associated sequences from which the Gaussian process can learn.

Figure \ref{fig:accuracy}(b) shows the classification accuracy (fraction of correct classifications), plotted against different training data sizes. The picture here is very different as compared to the balanced accuracy plot: it is clear that the algorithmic probability-based prior using $\hat{P}(x)$ of Eq.\ (\ref{eq:SB_pred}) improves upon the zero mean method for small training datasets of up to $\sim$3000 samples. While the absolute accuracy value is not very large, the relative value, when compared to $\sim$ 0.06 obtained from the zero mean method, is substantial.  Hence  the  algorithmic probability-based prior does offer tangible improvements over the zero mean benchmark (and baseline). See Appendix \ref{app:confusion} for more details on the classification performance, including some sample confusion matrices.

The explanation for this improvement is that the algorithmic probability-based prior `favours' the high probability classes (shapes), and because the testing data is made up of random sequences (simulating arbitrary typical sequences), accurate predictions of the high probability shapes have a large impact on the accuracy. In contrast, because there are only very few high-probability shapes, the increased classification accuracy of these few shapes has little impact on the balanced accuracy. 
Another reason why the accuracy improves substantially for larger training set sizes is that as the training set sizes increases, the training data composition becomes closer to a random collection of training sequences, which means that the true class probabilities are approximately given by the relative frequency of classes in the training data. Recall that for very small training set sizes, roughly equal numbers of training sequences are sampled from each class. But as the number of training samples increases, there are insufficient sequences in the $\sim$9000 training samples to allow for large numbers of training sequences for each of the 17 classes.

See Appendix \ref{app:L60} for classification accuracy and confusion matrix plots for the supplementary $L=$ 60 data. These plots show qualitatively similar patterns as compared to the $L=$ 100 plots, except that the improvement in accuracy is a little more pronounced.

\section{Discussion}

Motivated by exploring applications of algorithmic probability and algorithmic information theory (AIT) in machine learning (ML) problems, we here studied a realistic and practically relevant multiclass RNA molecule classification problem, using an algorithmic probability-based prior in a Gaussian process learning algorithm. Our main conclusion is that we found tangible classification accuracy score improvements in the low training-data regime. However, no substantial improvement in classification score was observed for balanced accuracy. To our knowledge, this work is one of the first empirical studies in which an algorithmic probability application has been made in a concrete, real-world ML problem. This is a promising start to an under-explored research interface between ML and AIT.

Our approach has some limitations, however. To begin, while the probability predictions for RNA shapes with $L=$ 100 were fairly accurate (and very accurate for $L=$ 60), it is often the case that less accurate probability predictions are made, or rather, only somewhat accurate upper bounds are predicted \cite{dingle2018input,alaskandarani2022low}. Hence, correspondingly poorer priors may be derived. Having said that, we suggest that a somewhat informative prior may be more useful than no prior at all, or equivalently,  a zero mean prior (but see \cite{owhadi2015brittleness} for cautions regarding priors). Additionally, having a prior which underestimates some class probabilities is perhaps not as problematic as overestimating class probabilities, because overestimating leads to the prior dominating and ignoring the information from the training data, while even the zero mean prior --- which gives a zero probability prior to each class --- works well.  

Another limitation is that the current approach is only relevant for classes that have some kind of pattern or shape, for which their complexity values can be estimated meaningfully (see also the discussion earlier in ref.\ \cite{hernandez2021algorithmic} on the problem of defining class complexities). Many real-world classification problems take the form of classes for which this complexity approach does not apply, e.g., Yes/No classes, or apple/orange/grape classes, which do not admit meaningful complexity estimates. Nonetheless, discrete shapes and patterns appear commonly in science and mathematics, and hence we expect that classes admitting meaningful complexity values will be found.

In future work, it would be interesting to explore other AIT-inspired predictions and algorithmic probability applications in other ML problems. Examples may include other classification problems, including adjusting the influence of the prior, possible applications in regression, time  series forecasting (e.g., \cite{dingle2022note}), and exploring regularisation which already uses the notion of complexity, and hence is a natural direction in which to incorporate information and complexity considerations. \\

\vspace{1cm}
\noindent
{\bf Acknowledgments:} 
This project has been partially supported by Gulf University for Science and Technology under project code: ISG --- Case grant number 253571 awarded to KD. We thank Fatme Ghaddar for providing the RNA sequences and shapes. We thank Boumediene Hamzi for discussions. HO and PBF gratefully acknowledge partial support by the Air Force Office of Scientific Research under MURI award number FA9550-20-1-0358 (Machine Learning and Physics-Based Modeling and Simulation).

\vspace{1cm}
\noindent
{\bf Data availability:} The data sets and code related to the
current study are available from the corresponding author on request.\\


\bibliographystyle{unsrt}
\bibliography{SBMLRefs} 

\onecolumngrid
\newpage
\appendix

\renewcommand\thefigure{\thesection.\arabic{figure}}

\section{Theory details}\label{usingAIT}

\subsection{Can we use AIT in practical settings, like machine learning and the natural sciences?}

AIT is an abstract theory developed in theoretical computer science. It is far from obvious that AIT should be applicable in real-world settings, like biology or ML. Indeed, the application of AIT to real-world problems suffers from several problems, including that (a) Kolmogorov complexity is uncomputable, (b) the results are framed and proved in the context of universal Turing machine (UTMs) while many real-world maps are not Turing complete, and (c) results are valid up to to $O(1)$ terms and therefore, strictly, only accurate in the asymptotic limit of large complexities. Given these, it is surprising that AIT can be successfully applied at all. 

The logic of the present study, as well as those of \cite{dingle2018input,dingle2020generic,johnston2022symmetry} (and others) is that we take AIT as a theoretical framework to make predictions and then test those predictions in practical settings, while being aware of the fact that strictly the predictions have not been proven to hold in the regime in which we apply them. We know that for asymptotically large patterns, the results from AIT apply, because we can ignore $O(1)$ terms in the asymptotic regime. Despite not being in the asymptotic regime, we use AIT results and empirically observe that they seem to work quite well, in the sense of making non-trivial predictions. 

It is not entirely clear to us why these AIT predictions work so well, and why we can `ignore’ these $O(1)$ terms, which could in principle dominate. However, ignoring these $O(1)$ terms is not specific to our work: in much of computer science, as well as often in applied mathematics, and also in physics, theory is developed or proven in asymptotic regimes  and then applied in practical, non-asymptotic, settings. In fact it is quite common to see results that are proved in the infinite limit, but work quite well far outside that regime. Why this is possible is an open question. Computer scientist Scott Aaronson \cite{aaronson2013philosophers} has pointed out that it is something of a mystery why merely examining the scaling form of equations is often good enough, and e.g.\  ignoring $O(1)$ terms is typically not a problem. Hence, even though there is some mystery in why our predictions work, just like in some areas of computer science, mathematics, and physics, this does not negate the usefulness of these predictions.

Despite not fully understanding the success of using AIT-inspired arguments, several lines of reasoning can help us understand why they are in fact applicable, at least approximately. 
Firstly, although technically uncomputable, Kolmogorov complexity is fundamentally merely a measure of the size in bits of the compressed version of a data object. Hence, in many situations, complexity can be approximated by standard compression algorithms (see below for more on this). Second, we the upper bound of ref. \cite{dingle2018input} is specifically relevant in the practical real-world (i.e.\ computable, non-universal Turing machine) setting. Relatedly, Calude et al.\   \cite{calude2011finite} have developed an AIT for finite state machines, suggesting that many fundamental ideas and results from AIT need not only apply to universal Turing machines.  Third, 
the inverse connection between probability and complexity is quite intuitive, and is not therefore only expected in the limited setting in which it is studied in AIT. 
Fourth, many of the fundamental ideas of AIT do not depend on UTMs and uncomputable quantities, but have analogs in computable settings. For example, the fundamental fact that most sequences are incompressible results from basic counting principles, and the fact that short programs appear with higher probability in many prefix codes holds outside of UTM settings. Fifth, because a UTM can simulate any computable function, we can always bound the behaviour of a computable function with known limits of the behaviour of a UTM. In addition, Kolmogorov complexity can be bounded by lossless compression algorithms. 

These points motivate using AIT as a theoretical framework to guide the derivation of mathematical predictions for real-world systems, which we might call AIT-inspired arguments. 





\subsection{Can we approximate Kolmogorov complexity?}

Kolmogorov complexity is an uncomputable quantity, meaning that it is not possible, even in principle, to calculate the complexity of a given string. The complexity can be bounded above, however. It is quite common in applications of AIT to use standard lossless compression algorithms to approximate Kolmogorov complexity \cite{li2008introduction}, such as gzip of Lempel-Ziv methods for compression. Clearly, these approximations have significant limitations, and cannot handle pseudo-random or chaotic patterns. See for example, Cosma Shalitz (\texttt{http://bactra.org/notebooks/cep-gzip.html}) on this.

On the other hand, although technically uncomputable, Kolmogorov complexity is fundamentally merely a measure of the size in bits of the compressed version of a data object. Hence, Vitanyi \cite{vitanyi2020incomputable,vitanyi2013similarity} points out that because naturally generated data is unlikely to contain pseudo-random complexities like $\pi$, the true complexity is unlikely to be much shorter than that achievable by every-day compressors. Therefore, how well standard compression algorithms work as approximations depends on the type of patterns that the system under study generates. If the system only has the computational capacity of a finite state transducer (or similar), then we can expect standard compressors to make reasonable complexity estimates. If the system has high computational capacity, or is expected to produce many pseudo-random patterns, then we can expect standard compressors not to make reasonable complexity estimates. For more on this discussion, see ref.\ \cite{vitanyi2020incomputable} and ref.\ \cite{teutsch2016brief} for work on short program estimates via short lists of candidates with short programs.

\subsection{Some conditions for observing simplicity bias}\label{ap:conditionsSB}

A full understanding of exactly which systems will, and will not, show simplicity bias (SB) is still lacking, but the phenomenon is expected to appear in a wide class of input-output maps, under fairly general conditions.

Some of these conditions were suggested in ref.\ \cite{dingle2018input}, including (1) that the number of inputs should be much larger than the number of outputs, (2) the number of outputs should be large, and (3) that the map should be `simple' (technically of $O(1)$ complexity) to prevent the map itself from dominating over inputs in defining output patterns. Indeed, if an arbitrarily complex map was permitted, outputs could have arbitrary complexities and probabilities, and thereby remove any connection between probability and complexity.  Finally (4), because many AIT applications rely on approximations of Kolmogorov complexity via standard lossless compression algorithms \cite{lempel1976complexity,ziv1977universal} (but see \cite{delahaye2012numerical,soler2014calculating} for a fundamentally different approach), another proposed condition is that the map should not generate pseudo-random outputs like $\pi=3.1415...$, which standard compressors cannot handle effectively. The presence of such outputs may yield high-probability outputs which appear `complex' hence apparently violating SB, but which are in fact simple. 

The ways in which SB differs from Levin's coding theorem include that it does not assume UTMs, uses approximations of complexities, and for many outputs $P(x)\ll 2^{-K(x)}$. Hence, the abundance of low-complexity, low-probability outputs \cite{dingle2020generic,alaskandarani2022low} is a signature of SB. Further, the upper bound was intended to apply to more general notions of complexity, such as the number of stems on an RNA molecule structure, not strictly Kolmogorov complexity.

\section{Methods}

 \subsection{Defining the classes and testing set}\label{app:classdefinition}
 
Defining the classes is not completely trivial in this RNA shapes problem. Because there are millions of possible shapes, most of which have exponentially small probabilities, larger and larger samples of random RNA sequences will keep finding more and more shapes.

In this study, we generated 10,000 RNA sequences, splitting them into $\sim$9000 for training and 1000 for testing. Within the $\sim$9000 training sequences, we found 17 unique shapes. Within the testing test, there were a few sequences whose shapes were not among these 17. In principle, we might have added these extra shapes to be included into the definition of the set of classes. However, this would imply that there are no associated training sequence for these one or two shapes. Hence, for the sake of simplicity, we defined the set of classes to be just the 17 found in the training data, and excluded those few sequences in the testing set. Hence the testing set had slightly less than 1000 sequences.

\subsection{Complexity}\label{app_complexity}

To approximate the Kolmogorov complexity we followed the methodology of ref.\ \cite{dingle2018input}, in which  the function
\begin{equation}
C_{LZ}(x) =\begin{cases}
     \log_2(n), &  \hspace*{-0.3cm}  \text{$x=0^n$ or $1^n$}\\
    \log_2(n) [N_w(x_1...x_n) + N_{w}(x_n...x_1)]/2, & \hspace*{-0.2cm} \text{otherwise}
  \end{cases}\label{eq:CLZ}
\end{equation}
was defined where $N_w$ forms the basis for the 1976 Lempel and Ziv  complexity measure  \cite{lempel1976complexity}. Here the simplest strings $0^n$ and $1^n$ are separated out because  $ N_{w}(x)$ assigns complexity $K=1$ to the string 0 or 1, but complexity 2 to $0^n$ or $1^n$ for $n\geq2$, whereas the true Kolmogorov complexity of such a trivial string actually scales as $\log_2(n)$ for typical $n$, because one only needs to encode $n$. 
The minimum possible value is $K(x)$$\approx$$0$ for a simple set, and so, e.g., for binary strings of length $n$, we can expect $0 \lesssim K(x) \lesssim n$ bits. Because for a random string of length $n$ the value $C_{LZ}(x)$ is often much larger than $n$, especially for short strings, we scale the complexity 
\begin{equation}
\tilde{K}(x) = \log_2(M) \cdot \frac{ C_{LZ}(x) - \min_x (C_{LZ})}{\max_x (C_{LZ}) - \min_x (C_{LZ}) } \label{eq:Kscaled}
\end{equation}
where $M$ is the maximum possible number of phenotypes in the system, and the minimum and maximum complexities are over all strings $x$ which the map can generate.  $\tilde{K}(x)$ is the approximation to Kolmogorov complexity that we use throughout. This scaling results in $0\leq \tilde{K}(x) \leq n$ which is the desirable range of values.

\subsection{Smoothing probabilities}\label{app_avprb}

Predicting probabilities using complexity estimates will yield, at best, roughly accurate predictions and are typically a little erratic.  Overestimating the probability of a given class will cause the resulting prior to classify all test sequences as belonging to that corresponding class, and ignore the test sequence itself. In order to reduce this, and reduce the effect of the erratic probability estimates, here we smooth the probabilities. This is done by taking the average of the (ranked) log probabilities, or equivalently the geometric mean of probabilities. 

Mathematically, this is given by

\begin{equation}
\hat{P}(x_i)  = 10^{\frac{ \log_{10}(2^{-\tilde{K}(x_{i-1})})  +   \log_{10}(2^{-\tilde{K}(x_i)}) + \log_{10}(2^{-\tilde{K}(x_{i+1})})    }{3}}
\end{equation}
for classes $x_1$, $x_2$, . . ., $x_C$ sorted in order of increasing probability, and $i=2,3,4,\dots,C-1$. For $x_1$ and $x_C$ the average is using only two values, namely $x_1$ and $x_2$, and $x_{C-1}$ and $x_{C}$. 
This arithmetic mean of log probabilities is equivalent to taking the geometric mean $(2^{-\tilde{K}(x_{i-1})}2^{-\tilde{K}(x_i)}2^{-\tilde{K}(x_{i+1})})^{1/3}$ for $i=2,3,4,\dots,C-1$.

\section{Length $L=$ 60 RNA}\label{app:L60}

In order to see the performance of the classification scores for a different data set, we here show plots for RNA of length $L=60$, as opposed to $L=100$ which is shown in the main text.  The methods followed are exactly the same as for the $L=100$ example: 10,000 random sequences were generated and split into a training set of $\sim$9000 sequences, and a testing set of 1000 sequences. For these $L=60$ data, there were 7 classes. The reason there are fewer shapes is that shorter sequences can form fewer possible shapes. 

\begin{figure}
\subfloat[]{\includegraphics[width=9cm]{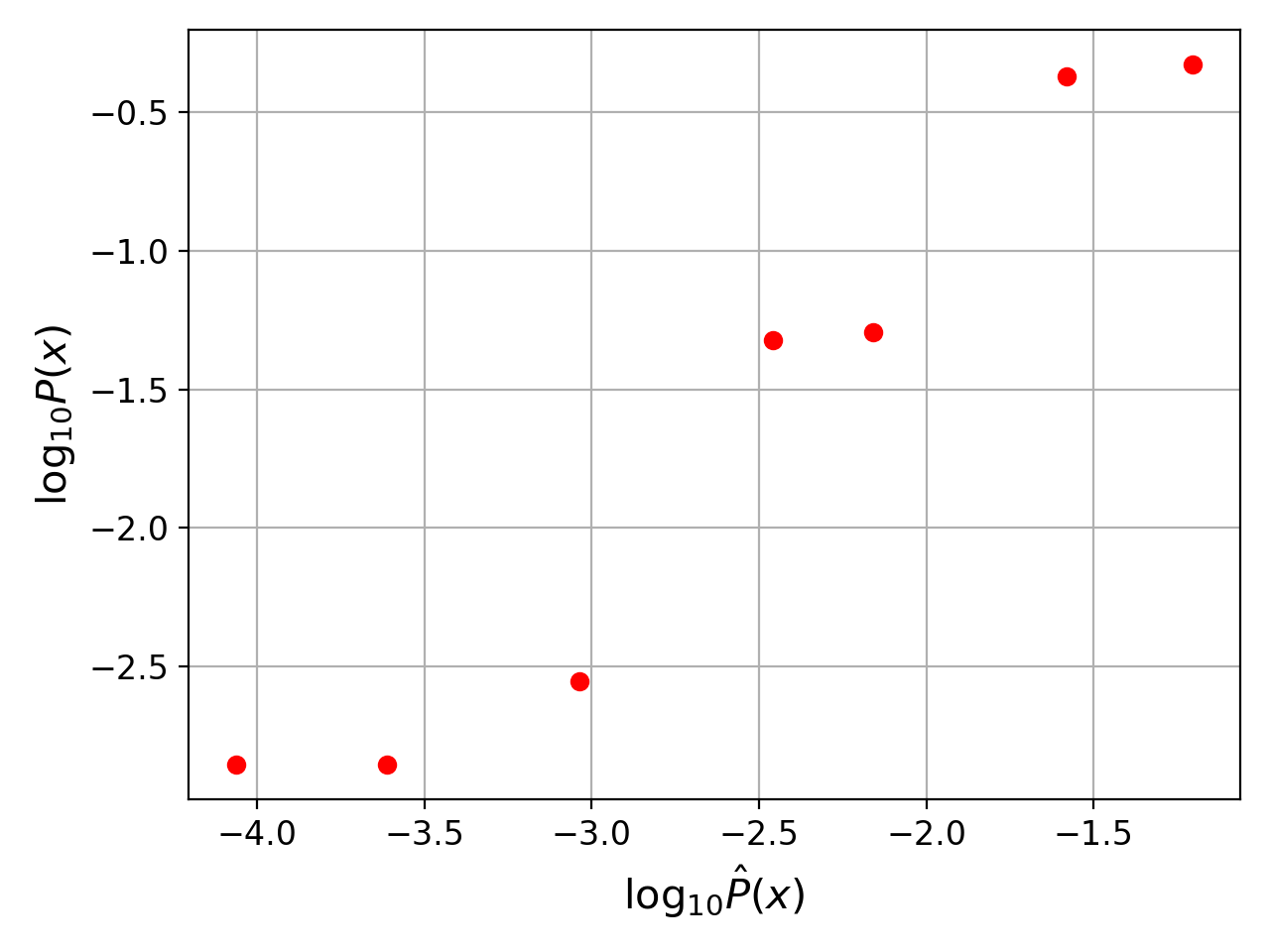}}
\caption{\textbf{Predicting class probabilities}. Predicted class probabilities  $\hat{P}(x)$ using Eq.\ (\ref{eq:SB_pred}) plotted against true class probabilities $P(x)$ for RNA shapes $L=60$. There are 7 shapes (classes). The linear correlation is very strong at $r=0.97$ ($p$-value = 0.0003).}
\label{fig:correlation60}
\end{figure}

Figure \ref{fig:correlation60} shows a correlation plot of the predicted and true class probabilities. The prediction accuracy is very high, with a linear correlation of $r=$ 0.97 and $p$-value$=$ 0.0003.

\begin{figure*}
\subfloat[]{\includegraphics[width=9cm]{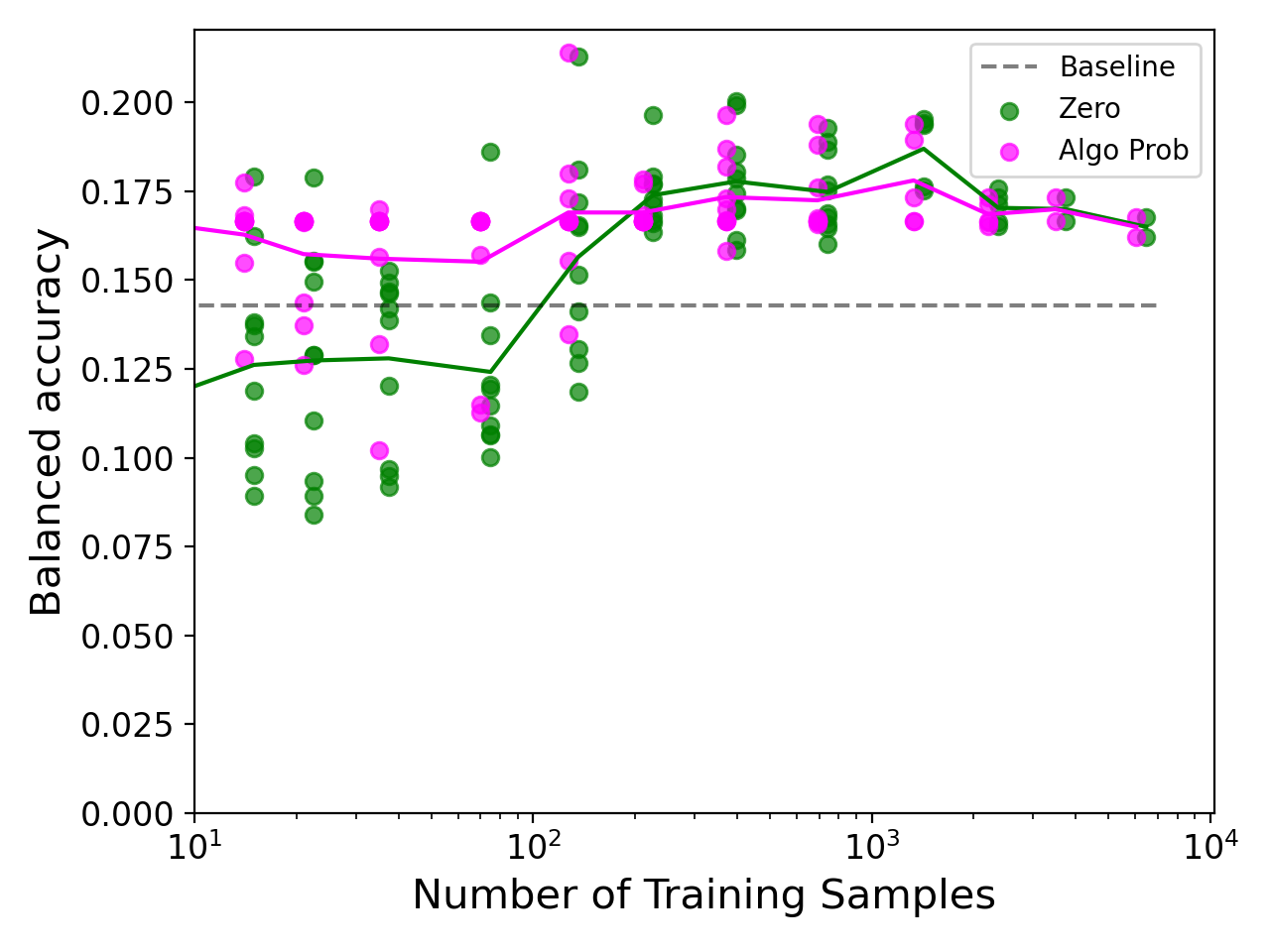}}
\subfloat[]{\includegraphics[width=9cm]{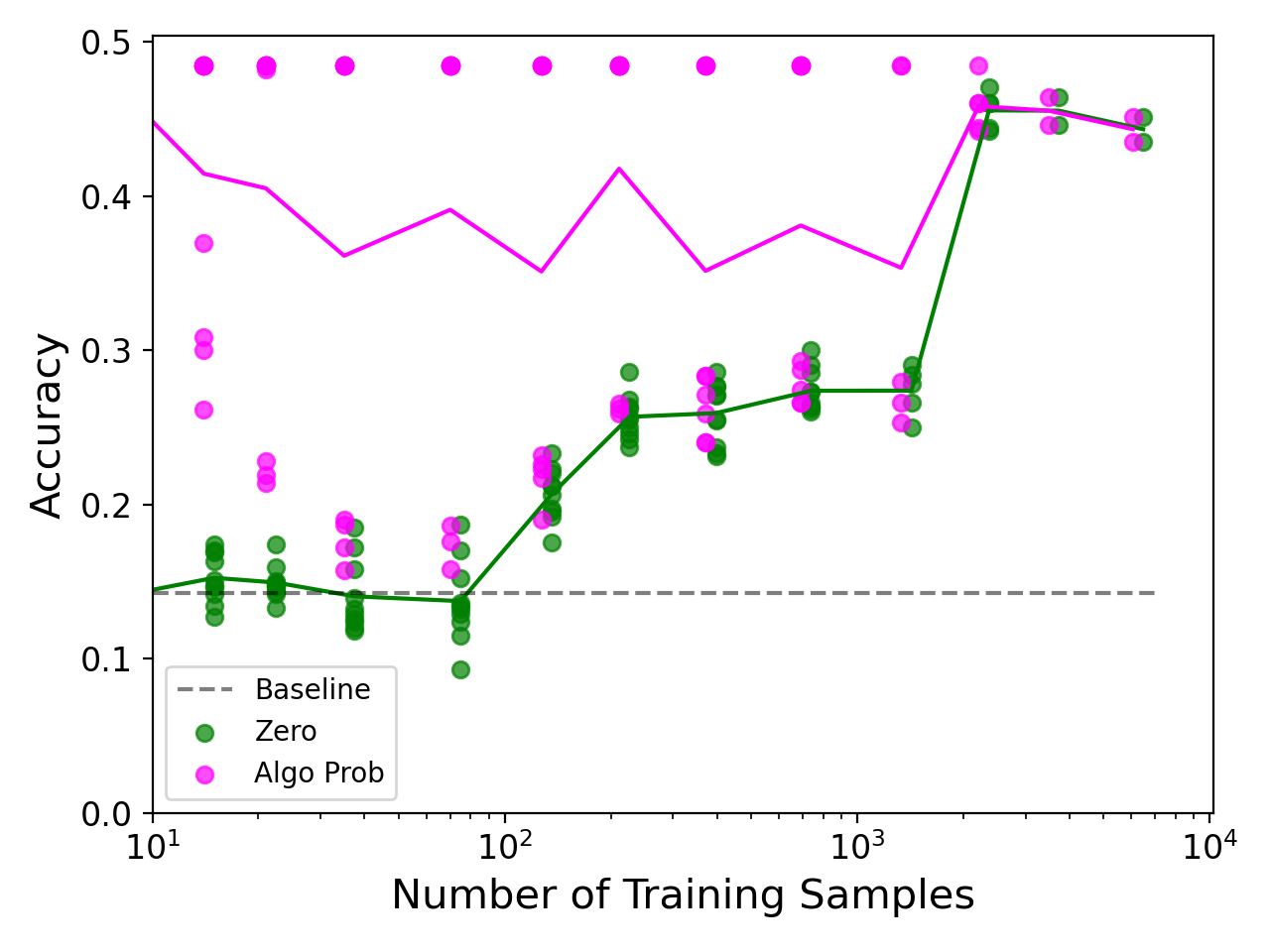}}\\
\caption{\textbf{Multiclass classification accuracy using different priors}. Comparing the accuracy for a 7-class supervised learning classification problem with a zero mean prior (green), and the algorithmic probability-based prior $\hat{P}(x)$ of Eq.\ (\ref{eq:SB_pred}) (pink) for different training sample sizes. Panel (a) shows mean balanced accuracy scores when equally weighting each class; (b) shows mean accuracy, i.e., simply the fraction of correct classifications. In (a), both priors yield similar accuracy scores, but in (b), the  $\hat{P}(x)$ prior outperforms the zero mean prior for smaller sample sizes. Note that a small multiplicative shift has been applied to separate the green and pink data, for the purposes of visualisation only.}
\label{fig:accuracy60}
\end{figure*}

Figure \ref{fig:accuracy60} shows the classification accuracy for these data. Similarly to the $L=100$ data, there is only a modest improvement in balanced accuracy scores depicted in panel (a). In contrast, the accuracy scores depicted in panel (b) are substantially higher for the algorithmic probability-based prior  than for the zero mean benchmark.

\begin{figure}
\subfloat[]{\includegraphics[width=9cm]{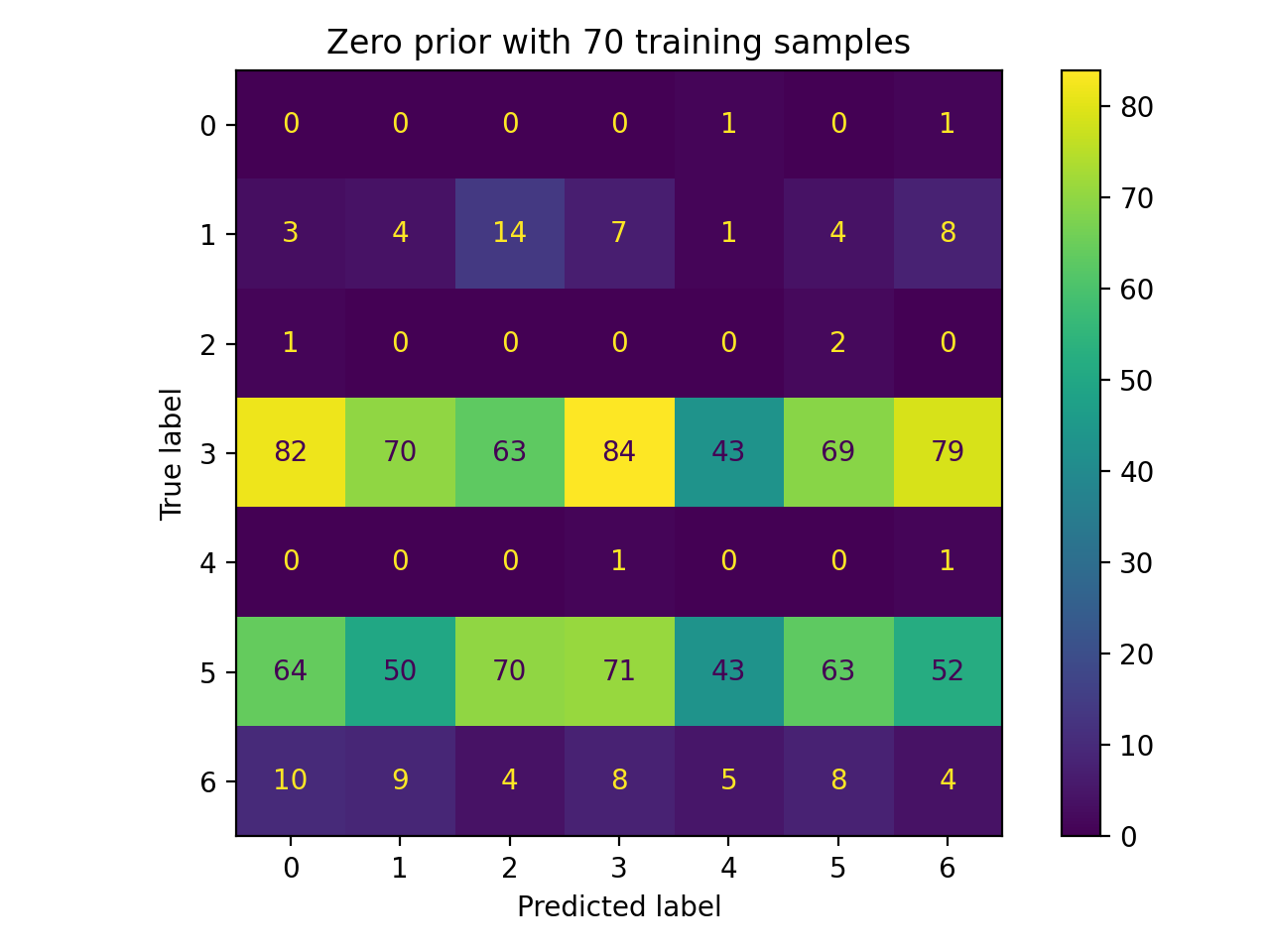}}
\subfloat[]{\includegraphics[width=9cm]{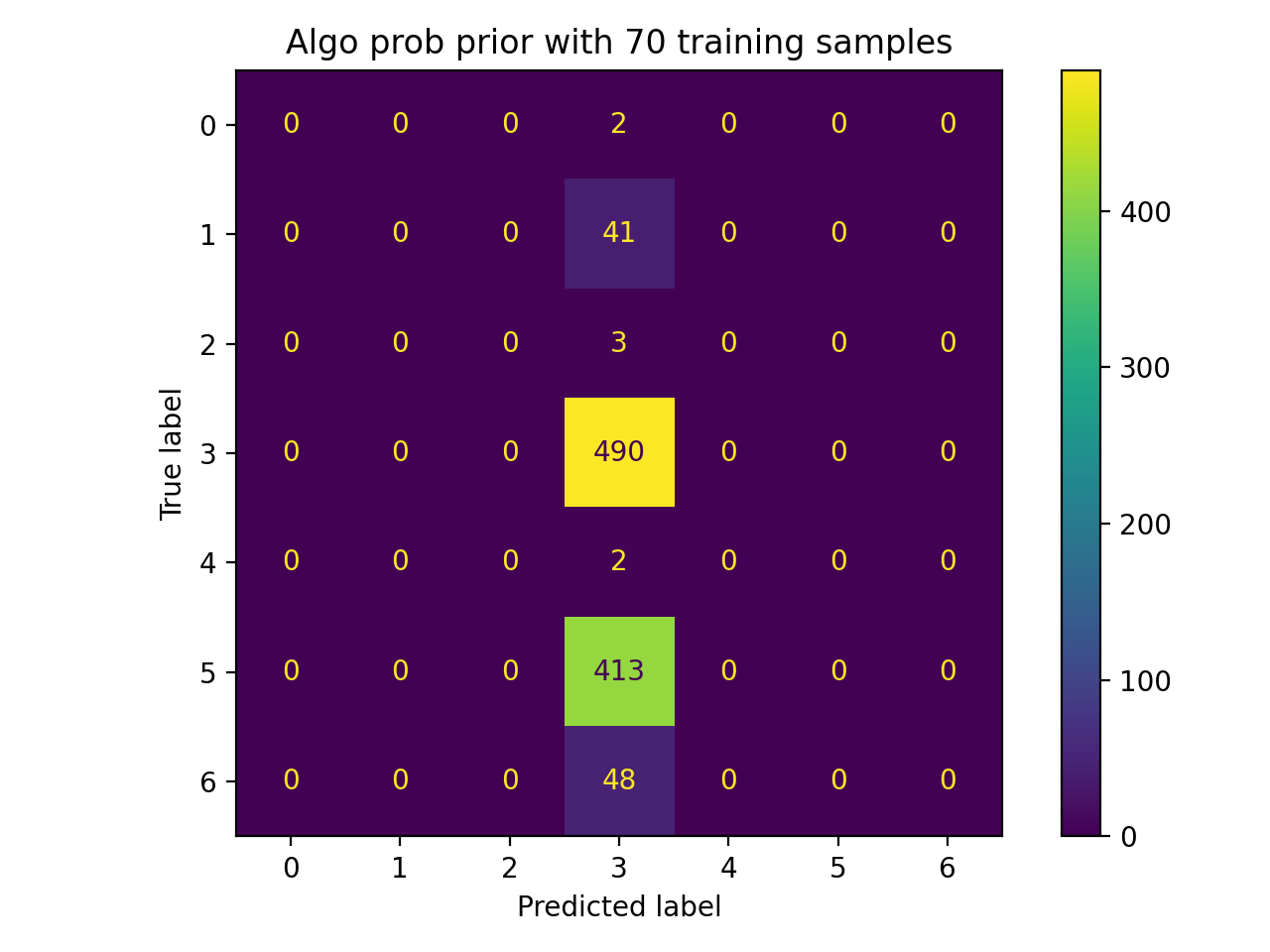}}\\
\subfloat[]{\includegraphics[width=9cm]{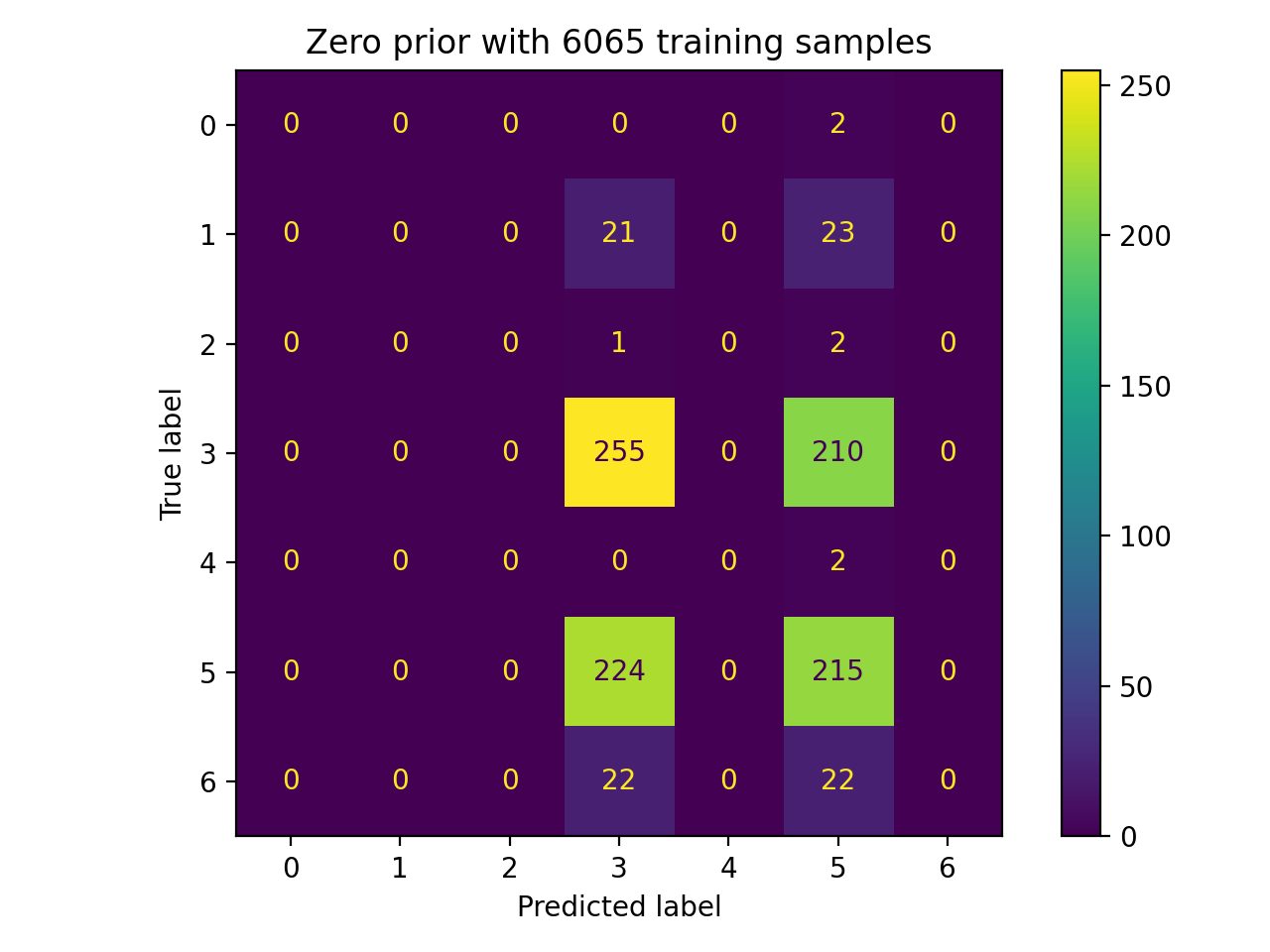}}
\subfloat[]{\includegraphics[width=9cm]{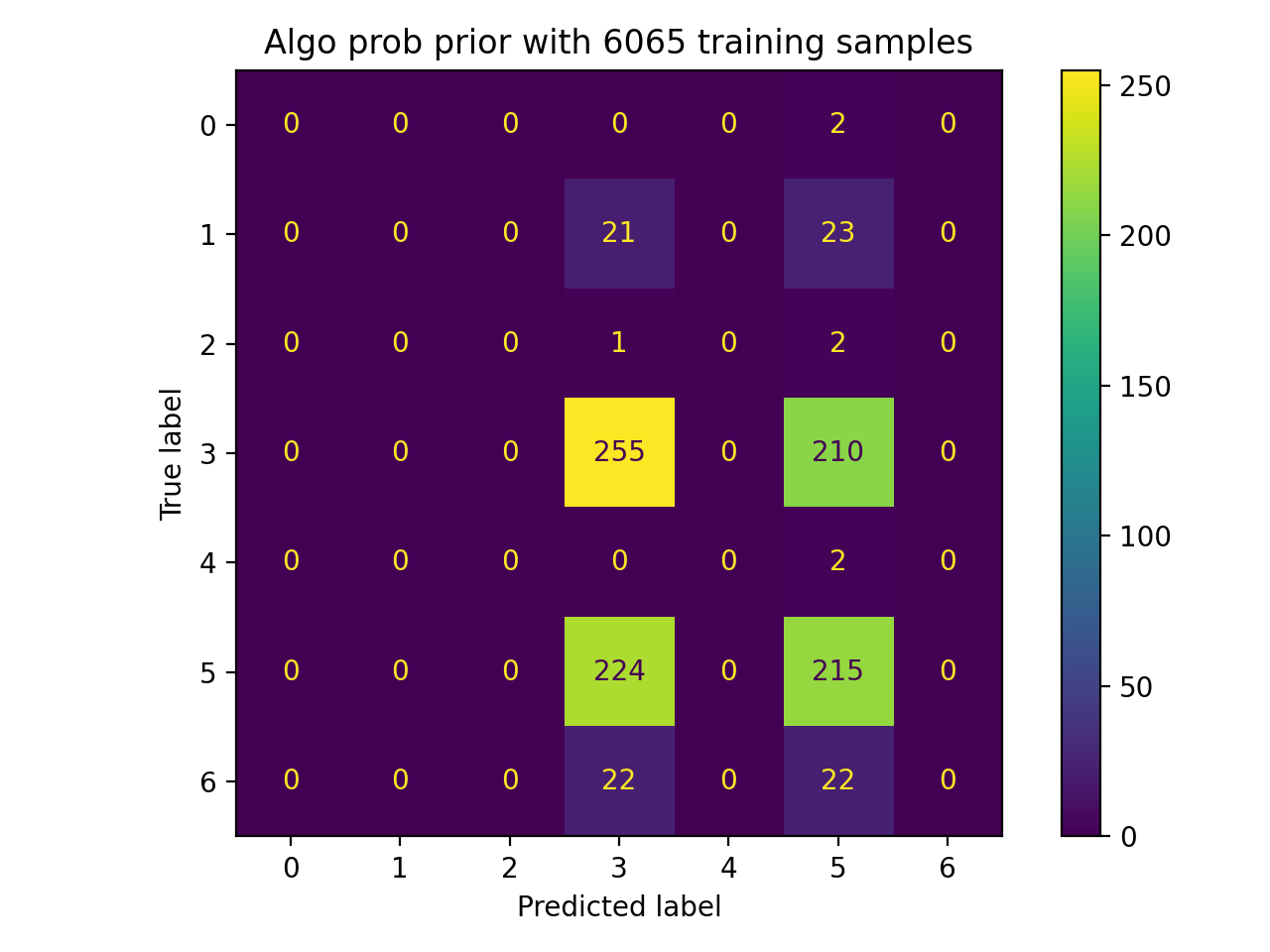}}
\caption{\textbf{Confusion matrices $L=60$}. Predicted classes vs true classes for the zero prior and algorithmic probability-based prior, for a very small sample size (70) and  larger training data sample size (6065) example.}
\label{fig:confusion60}
\end{figure}

Figure \ref{fig:confusion60} shows  sample confusion matrices. Again, these are qualitatively similar to those obtained from the $L=100$ data.

\section{Confusion matrices}\label{app:confusion}

In order to get a better insight into how the classification performance changes with increasing sample size, we plot confusion matrices for a very small sample size example (170 training samples), and a larger example (7133 samples). A confusion matrix shows the predicted and true classifications. Correct class predictions are counted on the diagonal of the matrix. Off-diagonals show incorrect classification counts.

\begin{figure}
\subfloat[]{\includegraphics[width=9cm]{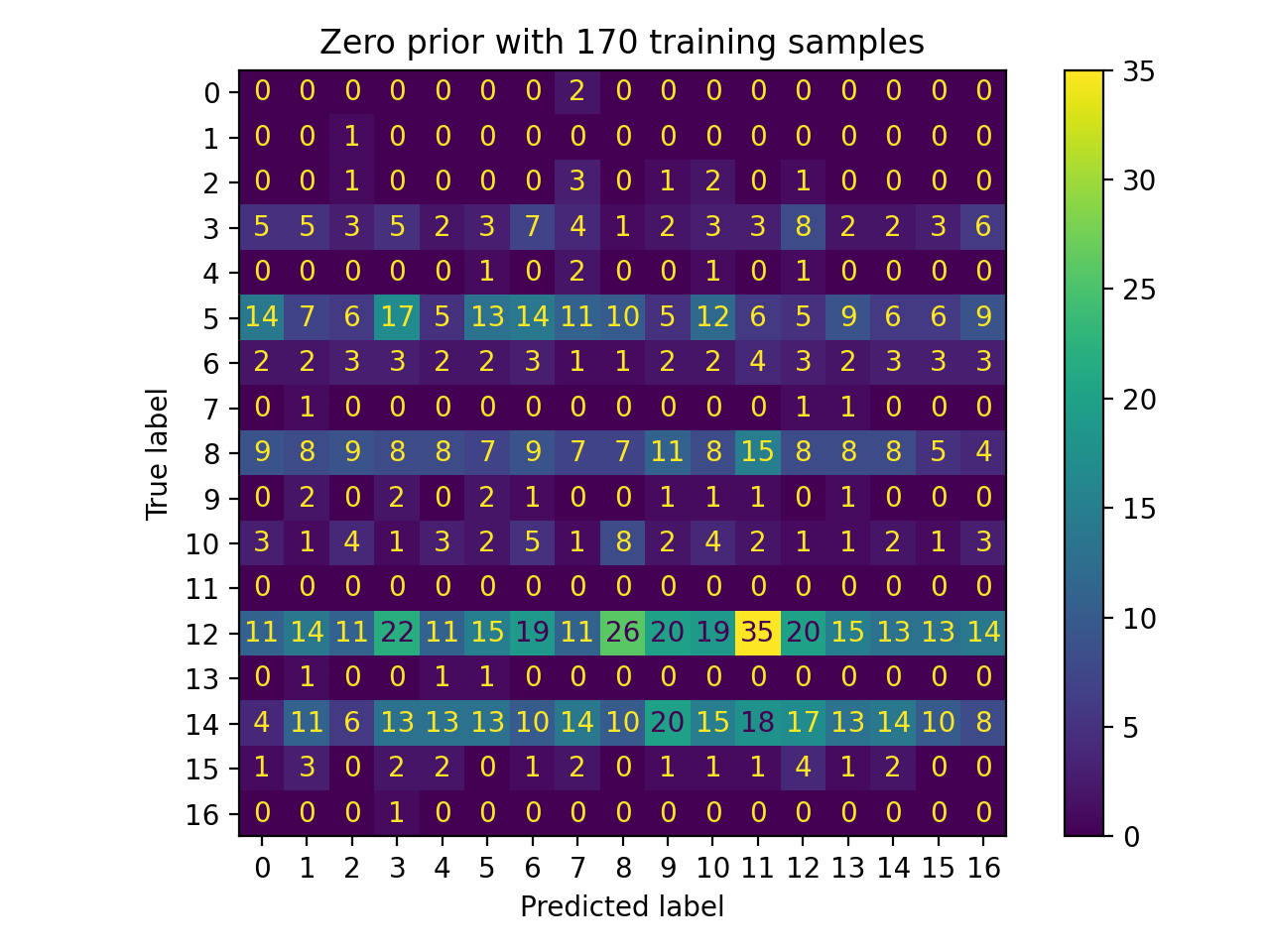}}
\subfloat[]{\includegraphics[width=9cm]{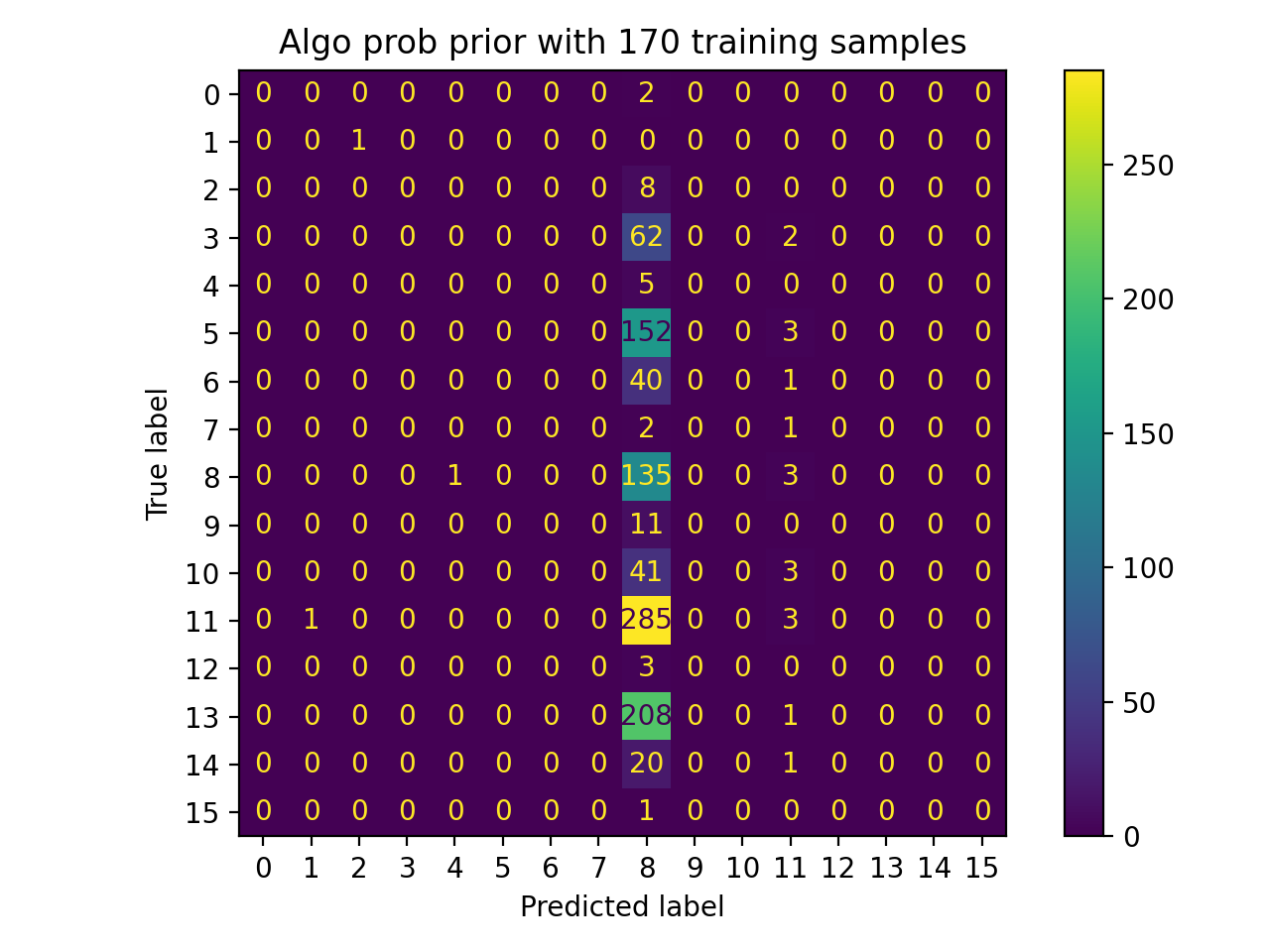}}\\
\subfloat[]{\includegraphics[width=9cm]{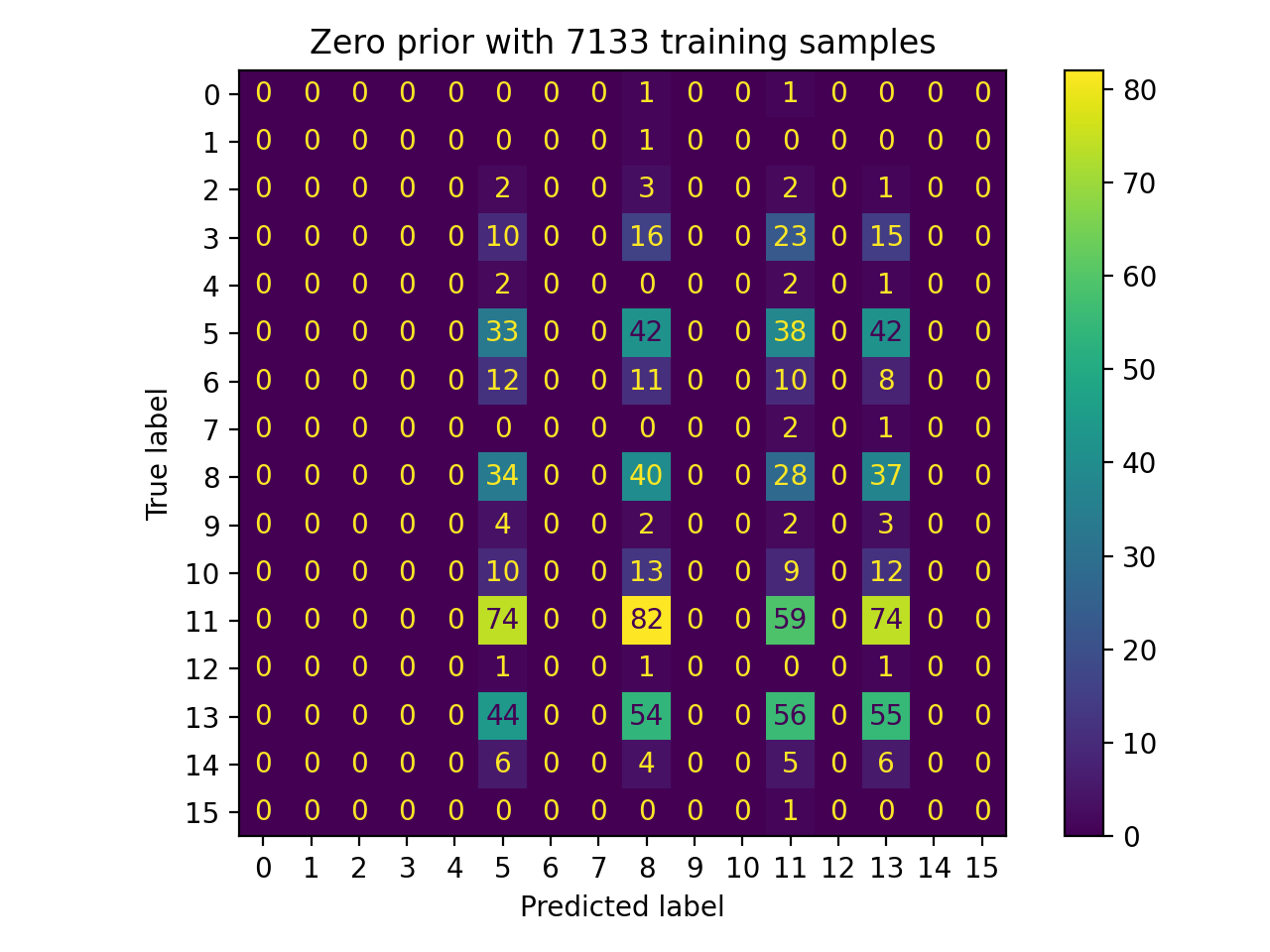}}
\subfloat[]{\includegraphics[width=9cm]{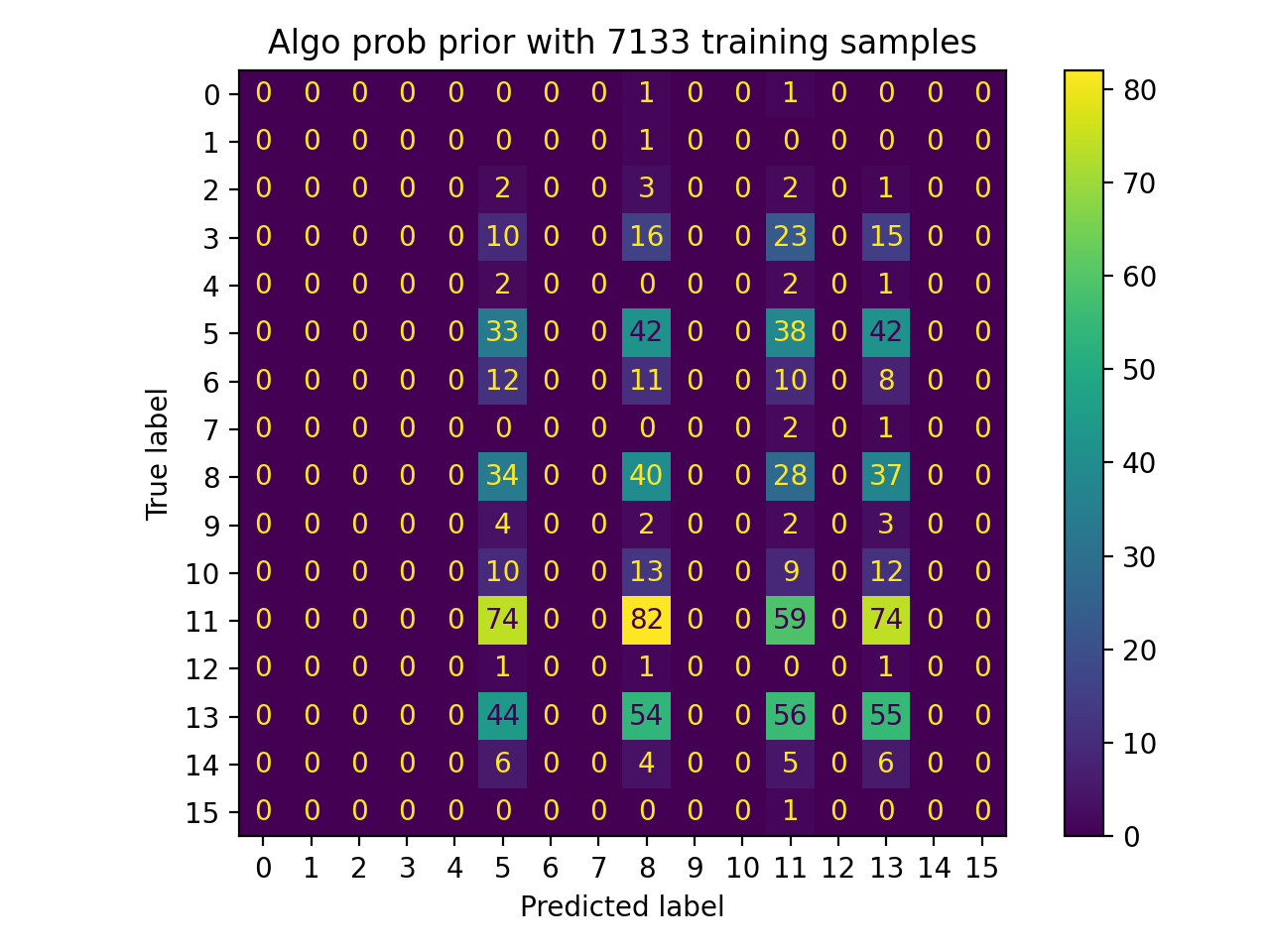}}
\caption{\textbf{Confusion matrices $L=100$}. Predicted classes vs true classes for the zero prior and algorithmic probability-based prior, for a very small sample size (170) and  larger training data sample size (7133) example.}
\label{fig:confusion}
\end{figure}

In Figure \ref{fig:confusion}(a), we see that the zero prior makes a fairly random range of predictions across many different classes, with very low accuracy. In Figure \ref{fig:confusion}(b) the matrix looks very different, with all testing samples being predicted to belong to only one class, the class with the highest $\hat{P}(x)$ probability. 
Figure \ref{fig:confusion}(c) and (d) show confusion matrices for a larger training sample size, namely 7133. In this case, the zero prior and algorithmic probability-based prior yield identical matrices. Both show that only four classes are predicted for each test sample. With increased training data, eventually, all of the classes would be predicted, not just the four with the highest $\hat{P}(x)$ probability.

The large difference between balanced accuracy and plain accuracy can be understood by examining  Figure \ref{fig:confusion}(c) or (d). We can see four classes that have a high probability, i.e., the classes labeled 5, 8, 11, and 13. For these, a non-trivial fraction of the predictions is correct (as gauged from the counts on the diagonals). On the other hand, the remaining 13 classes are never predicted at all, and the diagonals are all 0. Because 13 of the diagonal elements are 0 and hence of 0 accuracies, and only 4 have positive values, the balanced accuracy is very low. In contrast, most of the probability mass is associated with classes 5, 8, 11, and 13, and so the higher accuracy values for these classes contribute strongly.

Note that there are only 16 classes present in the matrices of Figure \ref{fig:confusion}, not 17. The reason for this is that the matrix is made up of the union of all classes
present in the testing set and those present in the predicted set. One of the 17 shapes with very low probability did not appear in this union and hence is not present in the matrix.

\end{document}